\documentclass[lettersize,journal]{IEEEtran}
\usepackage{array}
\usepackage{amsmath}
\usepackage{amsthm}
\usepackage{amssymb}
\usepackage{mathrsfs}
\usepackage{algorithm}
\usepackage{algorithmic}
\usepackage{amsfonts} 
\usepackage{booktabs}
\usepackage{float} 
\usepackage[pdftex]{graphicx}
\graphicspath{pics/}
\DeclareGraphicsExtensions{.pdf,.jpeg,.png,.jpg}
\usepackage{color}
\usepackage[dvipsnames, svgnames, x11names]{xcolor}
\usepackage{lipsum}
\usepackage{multirow} 
\usepackage{pifont}
\usepackage{soul}
\usepackage{subfigure}  
\usepackage{times}
\usepackage{url}
\usepackage{ulem}
\usepackage{verbatim}
\usepackage[hidelinks]{hyperref}

\usepackage{subcaption}
\usepackage{mathtools}
\usepackage{microtype}

\usepackage{fancyhdr}
\ifCLASSOPTIONcompsoc
  \usepackage[nocompress]{cite}
\else
  \usepackage{cite}
\fi
\ifCLASSINFOpdf
\else
\fi
\begin{document}
% \title{Anomaly Detection for 6G Networks via Federated Meta-Learning}
\title{Argus: Federated Non-convex Bilevel Learning over 6G Space-Air-Ground Integrated Network}

\author{Ya~Liu, 
        Kai~Yang, Yu~Zhu, Keying~Yang, Haibo~Zhao}%

\author{Ya~Liu, 
        Kai~Yang, ~\IEEEmembership{Senior Member,~IEEE}, Yu~Zhu, Keying~Yang, Haibo~Zhao %

% **待修改单位和邮箱
% \IEEEcompsocitemizethanks{\IEEEcompsocthanksitem Ya Liu, Kai Yang, Yu~Zhu, Keying~Yang, and Haibo~Zhao are with the Department of Computer Science and Technology, Tongji University, Shanghai 201804, China (e-mail: yaliu@tongji.edu.cn, kaiyang@tongji.edu.cn, 2331926@tongji.edu.cn, 1951859@tongji.edu.cn, 2052535@tongji.edu.cn). (Corresponding author: Kai Yang)}}% 

\IEEEcompsocitemizethanks{\IEEEcompsocthanksitem Ya Liu, Kai Yang, Yu~Zhu, and Keying~Yang are with the Department of Computer Science and Technology, Tongji University, Shanghai 201804, China (e-mail: yaliu@tongji.edu.cn, kaiyang@tongji.edu.cn, 2331926@tongji.edu.cn, 1951859@tongji.edu.cn). Haibo~Zhao are with the Department of Computer Science and Engineering, Hong Kong University of Science and Technology, Hong Kong, China (e-mail: hzhaobr@connect.ust.hk). (Corresponding author: Kai Yang)}}% 

% The paper headers
% \markboth{Journal of \LaTeX\ Class Files,~Vol.~14, No.~8, August~2015}{}%
%{Shell \MakeLowercase{\textit{et al.}}: Bare Demo of IEEEtran.cls for Computer Society Journals}

\IEEEtitleabstractindextext{%
\begin{abstract}
The space-air-ground integrated network (SAGIN) has recently emerged as a core element in the 6G networks. However, traditional centralized and synchronous optimization algorithms are unsuitable for SAGIN due to infrastructureless and time-varying environments. This paper aims to develop a novel \underline{A}synch\underline{r}ono\underline{us} al\underline{g}orithm a.k.a. Argus for tackling non-convex and non-smooth decentralized federated bilevel learning over SAGIN. The proposed algorithm allows networked agents (e.g. autonomous aerial vehicles) to tackle bilevel learning problems in time-varying networks asynchronously, thereby averting stragglers from impeding the overall training speed. We provide a theoretical analysis of the {iteration complexity}, communication complexity, and computational complexity of Argus. Its effectiveness is further demonstrated through numerical experiments.
\end{abstract}

% Space-air-ground Integrated Network (SAGIN) is expected to play an increasing role in providing real-time, flexible, and integrated communication and data transmission services in an efficient manner. 

\begin{IEEEkeywords}
% 关键词最多5个
bilevel learning, federated learning, decentralized optimization, 6G.
\end{IEEEkeywords}}

\maketitle

\IEEEdisplaynontitleabstractindextext

\IEEEpeerreviewmaketitle

\section{Introduction}
\label{sec:intro}
% 第一段 介绍sagin 发展、组成和优点。价值。
\IEEEPARstart{O}{ver} the years, with the implementation of the fifth-generation (5G) mobile communication technology, wireless communication technology has reached unprecedented heights \cite{boccardi2014five}. Despite advancements, the 5G network coverage is still insufficient, and the channel capacity of mobile communication remains limited. The terrestrial network segment may struggle to provide stable access to users in remote or non-terrestrial regions. To overcome these limitations, the space-air-ground integrated network (SAGIN) \cite{zhou2023aerospace} concept has been proposed for 6G and beyond. It integrates satellite networks in the space layer, air nodes (e.g., autonomous aerial vehicles etc.) in the air layer, and edge devices (e.g., Internet of Things devices) in the ground layer to facilitate seamless global connectivity \cite{fang2022olive}. Nowadays, SAGIN has been widely developed for position navigation, environment monitor, warning detection, disaster relief, etc. 
% Such integrated networks are now widely developed, with numerous countries, organizations, and scholars contributing plans and ideas \cite{shaengchart2024spacex}. 
%For instance, the Starlink project has launched over 6,000 satellites, providing coverage across 40 countries and supporting nearly 4 million users. 

% 第二段 sagin的应用，例如救灾等场景。
% 引出sagin中双层优化的必要性。
Due to the hierarchical structure and complex dependencies within SAGIN, bilevel optimization is more suitable for modeling optimization problems in this context compared to traditional single-level optimization. Bilevel optimization, also known as bilevel learning \cite{dempe2020bilevel}, addresses problems with a hierarchical structure, i.e., $\min_{\mathbf{x}} F(\mathbf{x}, \mathbf{y})$, s.t. $\mathbf{y} = \arg\min_{\mathbf{y}'} f(\mathbf{x},\mathbf{y}')$. It is especially suited for multi-tiered decision structures, allowing upper-level control of global objectives while lower levels solves specific tasks. Recently, bilevel optimization has been widely applied to various machine learning scenarios, such as meta-learning, reinforcement learning, and hyperparameter optimization. Distributed bilevel optimization has its roots in distributed robust optimization for communication networks \cite{yang2008distributed}, and has more recently gained attention in the context of machine learning such as federated learning and meta-learning\cite{jiao2022asynchronous}. Bilevel optimization’s ability to model interdependent variables offers a versatile framework for enhancing overall network performance. It shows strong potential for future applications in complex network systems, such as SAGIN. This optimization method has already been applied across various SAGIN domains, including in-orbit computation and communication \cite{ouyang2023joint}, wireless resource allocation \cite{chen2024robust, chen2025gpu}, covert communication \cite{feng2024covert}, and collaborative mission planning \cite{wang2024bilevel}.

% 第三段 引出收集数据和训练神经网络模型。集中式的缺点。联邦学习可以克服上述缺点，进行介绍。但传统的是集中式的，仍有缺陷。
As an integrated network, SAGIN transfers and processes massive data collected by entities in the hierarchical environment \cite{liu2023introduction}. A primary approach to harnessing the data in SAGIN is to aggregate this dispersed data on a cloud server for centralized training of machine learning or deep learning (DL) models. However, the cloud server is becoming a bottleneck for large-scale implementation due to data transmission overhead and rising privacy concerns around raw data exchange. To address these issues, a privacy-preserving distributed ML paradigm known as Federated Learning (FL) \cite{mcmahan2017communication} has been introduced. FL enables collaborative training of a shared global model by aggregating locally computed updates from devices, eliminating the need to transfer sensitive data. FL maintains data privacy and minimizes communication costs among distributed devices, making it efficient for SAGIN environments. 

% 第四段 集中式的缺点 引出去中心化，它如何更符合sagin的需求。现有的dfl做双层优化的工作
Conventional Federated Learning (CFL) relies on a central server to aggregate model updates from distributed devices, which can create communication bottlenecks and elevate the risk of a single point of failure \cite{beltran2023decentralized}. To address these challenges, decentralized federated learning (DFL) has emerged as a promising serverless architecture that connects multiple agents directly. By enabling agents to exchange model updates directly, DFL effectively reduces communication bottlenecks and mitigates the risks associated with centralization. Its scalability makes it particularly suitable for heterogeneous environments, such as those with varying bandwidth, latency, and data distribution \cite{yuan2022decentralized}, which meets the demands of dynamic, large-scale data processing required in SAGIN. Recently, several decentralized bilevel optimization algorithms \cite{liu2022interact, chen2024decentralized, lu2022stochastic, lu2022decentralized, yang2022decentralized, liu2023prometheus, gao2023convergence, chen2023decentralized} have been proposed to support multi-agent and even multi-task learning, further advancing DFL's application potential. 

The aforementioned works rely on static network topologies, while dynamic topology modeling more accurately captures the true dynamics of SAGIN. However, frequent changes in topology can disrupt the consistency of information transmission. Moreover, these works typically assume homogeneous agents capable of synchronous iterations. In contrast, SAGIN nodes are highly heterogeneous, with significant differences in computational and communication capabilities. Synchronous communication mechanisms are greatly hindered by stragglers \cite{ozfatura2020straggler}. To enhance system efficiency, asynchronous algorithms must be designed to accommodate heterogeneous agents. However, this introduces more complex communication mechanisms, further exacerbating information inconsistency. Overall, the dynamic and heterogeneous nature of SAGIN significantly complicates system modeling and presents substantial challenges in ensuring algorithm convergence. Additionally, in the context of optimization, even solving linear bilevel optimization problems is NP-hard, and the solution to non-convex, non-smooth bilevel optimization problems in decentralized settings remains underexplored. In this common yet complex problem setup, the complexity of solving each problem layer is high, and the decentralized environment further complicates the coupling between the bilevel optimization problems.

% 第六段  我们做了什么，它呼应解决了sagin中的什么问题
% 后面在正文应该引出虚拟节点的概念
To address these issues, this paper introduces an \underline{A}synch\underline{r}ono\underline{us} al\underline{g}orithm (named Argus)  for federated bilevel optimization over SAGIN, where heterogeneous nodes dynamically cooperate to solve complex optimization problems. Argus adopts a decentralized structure to facilitate privacy-preserving model sharing across nodes, effectively mitigating common bottlenecks associated with central servers. To accommodate SAGIN's dynamic and heterogeneous nature, we implement an asynchronous communication approach in Argus, enabling nodes to train local models independently, without waiting for delayed nodes. Moreover, we target bilevel optimization problems that are more complex than those explored in prior studies. Specifically, we examine bilevel problems with non-smooth, non-convex objectives and consensus constraints at both levels, diverging from existing research that primarily addresses strongly convex lower-level objectives \cite{chen2024decentralized, lu2022stochastic, liu2023prometheus, gao2023convergence}. To solve these problems efficiently, we present a decentralized cutting planes algorithm, which approximates and addresses the bilevel optimization challenges. While finding a global minimum in non-convex settings may be NP-hard or even NP-complete \cite{murty1985some}, the solution can be considered exact as the number of cutting planes approaches infinity \cite{schrijver1980cutting}.

\textbf{Contribution.} The main contributions of our paper are summarized as follows: 
\begin{itemize}
    \item Different from existing decentralized bilevel optimization approaches that adopt static networks and synchronous communication mechanisms, Argus solves decentralized federated bilevel learning in a time-varying network asynchronously, enabling fast training in the infrastructureless environment. To the best of our knowledge, this paper represents the first attempt to tackle federated bilevel learning over SAGIN. 
    \item We establish convergence guarantees for Argus with non-convex and non-smooth objectives under consensus constraints of both upper-level and lower-level problems. Theoretical results indicate that Argus enjoys an {iteration complexity} of $\mathcal{O}(1/\epsilon)$ to achieve an $\epsilon$-stationary point. Additionally, we derive the communication and computational complexity of the proposed algorithm.
    \item Extensive experiments on multiple tasks and a range of public datasets have demonstrated the effect of Argus, which could outperform state-of-the-art baselines. 
\end{itemize}

% 论文结构
The remainder of this paper is organized as follows. In Section \ref{sec:relatedwork}, we summarize the related work. Section \ref{systemmodel} presents the system model and the proposed Argus algorithm is expounded in Section \ref{Argus}. Section \ref{discussion} gives theoretical analysis of the {iteration complexity}, communication complexity, and computational complexity of Argus. Section \ref{experiments} conducts the performance evaluation. The paper is concluded in Section \ref{conclusion}. Finally, we discuss the future directions in Section \ref{future}.

% \begin{figure}[htbp]
% \centering
% \includegraphics[width=0.7\linewidth]{fig/star_fig1.pdf}
% \caption{Illustration of SAGIN.} %  %大图名称
% \label{fig:star_fig1}  %图片引用标记
% \end{figure}

% \begin{figure*}[htbp]
% \centering
% \includegraphics[width=0.8\linewidth]{fig/star_fig234.pdf}
% \caption{Illustration of Argus in a time-varying network with $N = 9$ nodes.} %  %大图名称
% \label{fig:star_fig2}  %图片引用标记
% \end{figure*}

\begin{table*}[htb]%[!t]
\caption{A comparison of Argus with related prior works on decentralized bilevel optimization learning.}
\label{table1}
\centering
\begin{tabular}{c|c|c|c|c|c|c}
\hline 
\hline 
\multirow{2}{*}{\centering Work} & Upper-Level (UL) & Lower-Level (LL) & \centering Consensus & \multirow{2}{*}{\centering Non-Smooth} & \multirow{2}{*}{\centering Asynchronous } & Dynamic\\
&  Functions &  Functions & Constriants & & & Networks\\ 
\hline 
\cite{liu2022interact} & Non-Convex &
Strongly-Convex & UL & $\times$ & $\times$ & $\times$\\
\cite{chen2024decentralized} & Non-Convex & Strongly-Convex & $\times$ & $\times$ & $\times$ & $\times$\\
\cite{yang2022decentralized} & Non-Convex &
Strongly-Convex & $\times$ & $\times$ & $\times$ & $\times$\\
\cite{lu2022decentralized} & Non-Convex &
Strongly-Convex & UL & $\times$ & $\times$ & $\times$\\
\cite{lu2022stochastic} & Non-Convex &
Strongly-Convex & UL \& LL & $\times$ & $\times$ & $\times$\\
\cite{qiu2023diamond} & Non-Convex &
Strongly-Convex & UL & $\times$ & $\times$ & $\times$\\
\cite{chen2023decentralized} & Non-Convex &
Strongly-Convex & $\times$ & $\times$ & $\times$ & $\times$\\
\cite{gao2023convergence} & Non-Convex &
Strongly-Convex & $\times$ & $\times$ & $\times$ & $\times$\\
\cite{liu2023prometheus} & Non-Convex &
Strongly-Convex & UL & UL & $\times$ & $\times$\\
\hline
Argus & Non-Convex & Non-Convex & UL \& LL & $\checkmark$ & $\checkmark$ & $\checkmark$\\
\hline 
\end{tabular}
\end{table*}

\section{Related Work} 
% ** 补充这些主题在sagin下的工作
\label{sec:relatedwork}
In this section, we begin by discussing the characteristics, modeling, and solution methods for bilevel optimization problems (see Section II.A). Next, we examine the current research landscape on decentralized federated learning and review a range of studies that address bilevel optimization problems within the context of decentralized federated learning (see Section II.B). Finally, we present key ideas and recent developments related to solving optimization problems through polytope approximation. For clarity, Table I provides a summary of the latest related works, emphasizing their primary characteristics and comparing them with this paper.

\subsection{Bilevel Optimization}
% 这里的应用都是求解具体sagin问题的，我们是关注在sagin背景下广泛的机器学习任务。

% 形式化介绍
Bilevel optimization is a specialized hierarchical optimization problem \cite{dempe2020bilevel}, where the objective in the upper-level is optimized subject to constraints derived from the optimal solution of a lower-level optimization problem, and it has attracted significant research attention in recent years. A classical mathematical formulation of this problem is given by: $\min_{\mathbf{x}} F(\mathbf{x}, \mathbf{y})$, s.t. $\mathbf{y} = \arg\min_{\mathbf{y}'} f(\mathbf{x},\mathbf{y}')$, where $\mathbf{x}$ and $\mathbf{y}$ represent the upper- and lower-level variables, respectively, and $F(\mathbf{x}, \mathbf{y})$ and $f(\mathbf{x}, \mathbf{y}')$ denote the objective functions at the upper and lower levels. In addition, both the upper-level and lower-level optimization problems may each have their own additional constraints. In this formulation, the lower-level problem serves as a constraint, defining the relationship between $\mathbf{x}$ and $\mathbf{y}$, while the upper-level problem constitutes the main optimization task.

% 求解 先说双层优化的方法 
One strategy for addressing such problems involves gradient-based bi-level optimization, where the lower-level variables $\mathbf{y}$ can be estimated using gradient descent, and the upper-level optimization problem computes the hypergradient $\frac{d F(\mathbf{x}, \mathbf{y})}{d \mathbf{x}} = \frac{\partial F(\mathbf{x}, \mathbf{y})}{\partial \mathbf{y}'} \frac{\partial \mathbf{y}'}{\partial \mathbf{x}} + \frac{\partial F(\mathbf{x}, \mathbf{y})}{\partial \mathbf{x}}$ to update $\mathbf{x}$ \cite{liao2018reviving}. However, the computation of the hypergradient typically involves high-order complexity due to second-order derivatives, making exact computation challenging. Depending on the different approaches for approximating the hypergradient, the gradient-based bi-level optimization approaches can be further categorized into approximate implicit differentiation (AID) based approaches \cite{lorraine2020optimizing} and iterative differentiation (ITD) based approaches \cite{grazzi2020iteration}. The AID approaches avoid the direct computation of second-order derivatives by approximating implicit differentiation, while the ITD approaches approximate the derivatives through an iterative process.

%然后是转化为单层的方法
In contrast to the gradient-based methods, which require the computation of high-order derivatives and are computationally expensive, converting bilevel optimization problems into single-level optimization problems offers significant advantages. This approach eliminates the need for hypergradient computation by expressing the solution of the inner-level problem as a function of the upper-level variables. Consequently, the complex nested structure of bilevel problems is reduced, enabling the use of standard single-level optimization techniques. Typical methods for transforming bilevel problems into single-level ones include replacing the lower-level optimization problem with its analytical solutions \cite{zhang2022revisiting} or Karush-Kuhn-Tucker (KKT)\cite{biswas2019literature} conditions. However, assuming the lower-level optimization problem to have an analytical solution is often impractical. Besides, the resulting problems after the KKT conditions replacement could involve lots of constraints.

% 应用，并转折到去中心化
Bilevel optimization has found many applications in SAGIN, such as in-orbit computation and communication \cite{ouyang2023joint}, wireless resource allocation\cite{sun2022learning}, covert communication\cite{feng2024covert} and collaborative mission planning\cite{wang2024bilevel}. However, most of the existing approaches focus on centralized settings. These approaches may face data privacy risks \cite{subramanya2021centralized} since they require collecting data from large amount of distributed devices. Besides, centralized distributed bilevel optimization methods \cite{ji2021bilevel} use the parameter-server architecture to avoid the data privacy risks, but the communication bottleneck of the server hinders the practical application of these approaches.

\subsection{Decentralized Federated Learning}
% 联邦学习架构介绍
Federated Learning (FL) \cite{mcmahan2017communication} is a privacy preserving distributed learning paradigm that learns a shared model by aggregating locally-computed updates on multiple devices. The potential of the FL framework in SAGINs has been extensively explored in applications such as traffic anomaly detection \cite{xu2023anomaly}, intelligent transmission \cite{tang2022federated}, and traffic offloading \cite{qin2024differentiated}. However, these studies utilize a traditional centralized federated learning (CFL) framework, comprising a central server and client devices. In this setup, clients train local models on private data and periodically send model updates to the server. The server aggregates these updates to form a global model, which is then redistributed to clients for further training. This process repeats until convergence.

% 去中心化的优势
Due to the limited communication resources and unstable channels of SAGIN, the CFL framework is susceptible to communication bottlenecks and single-point failures at the server side. In a decentralized federated learning (DFL) framework \cite{sun2021decentralized}, the data is distributed among multiple agents, each of which communicates with its neighbors to solve a finite-sum minimization problem. DFL aims to solve problems without relying on a central server to collect iterates from local agents, and it enhances robustness to low network bandwidth compared to CFL by promoting decentralized aggregation \cite{beltran2023decentralized}. The primary challenge in the decentralized setting is the data heterogeneity among agents, which can be alleviated through communication \cite{chen2024decentralized}.

% 近期具体工作介绍
Recently, there have been works considering bilevel optimization under the DFL setting \cite{liu2022interact, chen2024decentralized, lu2022stochastic, lu2022decentralized, yang2022decentralized, liu2023prometheus, gao2023convergence, chen2023decentralized}. The main characteristics of these works are summarized in Table \ref{table1}, highlighting the comparison in terms of problem characteristics and network models. First, existing studies focus solely on strongly convex lower-level problems, which are easier to solve but have limited practical relevance, as many deep learning-related scenarios require the study of non-convex problems. The main challenges of solving non-convex problems include the difficulty of finding global optima and the potential for local minima, which complicates the optimization process. Furthermore, some works do not account for constraints in the optimization problems or only consider constraints in the upper-level optimization, limiting their generalizability. Constraints introduce additional complexity to the optimization structure. Moreover, most existing works focus on smooth optimization problems, but non-smooth problems often arise when dealing with non-differentiable loss functions. In Prometheus \cite{liu2023prometheus}, a decentralized algorithm based on proximal tracked stochastic recursive estimators is proposed for a nonconvex-strongly-convex bilevel problem with constraints in the non-smooth upper-level problem. Finally, existing works generally assume static and homogeneous networks, which are difficult to directly apply in the highly time-varying and device-heterogeneous environment of SAGIN. As shown in Table \ref{table1}, existing works rely on synchronous communication mechanisms, which are inefficient in SAGIN environments with significant device heterogeneity. A main drawback of synchronous DFL is its slow training speed due to waiting for stragglers. In contrast, asynchronous DFL has a faster training speed since each agent updates independently, but analyzing asynchronous algorithms theoretically is more challenging \cite{jiao2022asynchronous}. 

{Several asynchronous federated learning algorithms have been proposed. In the CFL setting, methods like FedAsync \cite{xie2019asynchronous} and EAFL \cite{zhou2024towards} adopt fully asynchronous approaches, where the central server promptly aggregates the global model upon receiving a local update. The semi-asynchronous method proposed in \cite{kou2024semi} allows the aggregation server to store earlier-received local models and aggregate them within a specified time frame. In the DFL setting, which avoids central bottlenecks, asynchronous algorithms such as AsyDFL \cite{liao2024asynchronous} and AEDFL \cite{liu2024aedfl} require each agent to communicate gradients only with a subset of neighbors to improve resource efficiency. In \cite{xie2024decentralized}, each agent receives stale parameters from all its neighbors within a given time window. The setup most similar to ours is found in \cite{jeong2022asynchronous}, where a subset of agents completes iterations within a fixed time frame. However, most existing asynchronous decentralized approaches focus on single-level optimization, while the complex interdependencies in bilevel optimization present additional challenges for asynchronous algorithm design.}

\subsection{Polyhedral Approximation}
% 基础定义
In optimization, polyhedral approximation refers to the technique of approximating a complex set or feasible region by using  a simpler or more tractable polyhedra. The main concept behind them is to construct a polyhedral outer approximation of the feasible set, and use the approximation to form a linear relaxation of the target problem \cite{lundell2022polyhedral,NLP-decode}. This strategy can simplify the optimization problem, reduce the computational complexity, and still obtain meaningful results. 

% 割平面法介绍
Cutting plane \cite{yang2014distributed} (or outer linearization) is one of the classical polyhedral approximation methods. The cutting planes are designed to exclude parts of the feasible region that do not contain optimal solutions, thereby refining the outer approximation of the feasible region. As the cutting planes are added, they progressively improve the approximation, bringing the feasible region closer to the original set and guiding the solution toward optimality. Furthermore, this exclusion of non-optimal solutions helps significantly reduce the problem space and accelerate convergence. Cutting plane methods have proven to be a powerful tool for solving complex optimization problems, performing well in areas such as integer programming, mixed-integer programming, and non-convex optimization. This method is flexible and can be customized to construct cutting planes based on the specific characteristics of the problem at hand. Additionally, cutting plane methods exhibit a high degree of decomposability, which enhances their scalability and makes them particularly well-suited for distributed scenarios.

The advantages of the cutting plane method mentioned above have attracted researchers to apply it to solving complex optimization problems such as bilevel optimization \cite{kleinert2021survey, cao2024projection, jiao2024provably}. However, these methods are still limited to centralized algorithms. In distributed settings \cite{distributed-wireless,distributed-sgrid,distributed-sgrid2}, a core challenge lies in how to appropriately decompose the problem into several subproblems that can be solved locally, as well as designing communication mechanisms for cutting plane parameters. \cite{yang2008distributed,jiao2022distributed,jiao2022asynchronous} proposed cutting plane methods for distributed bilevel optimization problems and provided convergence proofs, demonstrating the feasibility and effectiveness of the cutting plane method for distributed solutions of non-convex bilevel optimization problems. However, these works adopt a centralized federated learning framework, utilizing a single server node to construct the cutting planes. In decentralized settings, the process of cutting plane construction and problem-solving becomes more complex and has not been sufficiently explored. Moreover, existing works only consider smooth objective functions, limiting the ability to optimize non-smooth objective functions.

% 待加入SYSTEM MODEL 
\section{SYSTEM MODEL}
\label{systemmodel}

% 需要补充一个图
\subsection{Network Model}
\label{networkmodel}
\begin{figure*}[htbp]
\centering
\includegraphics[width=0.7\linewidth]{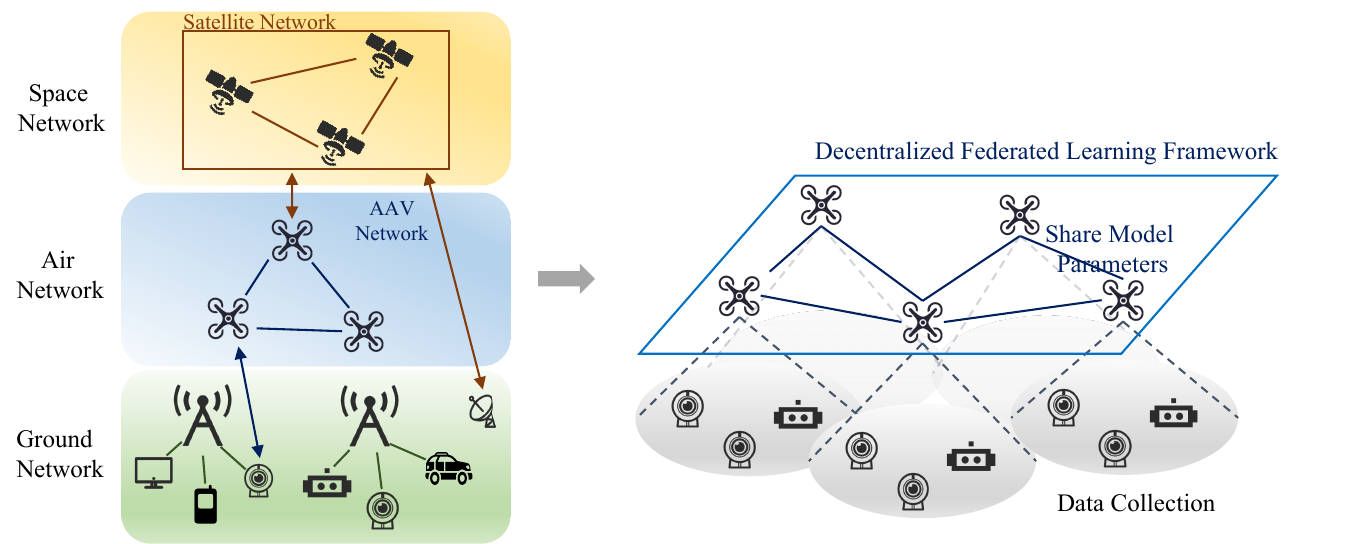}
\caption{The architecture of decentralized federated learning framework in SAGIN.} %  %大图名称
\label{fig:sagin}  %图片引用标记
\end{figure*}

% 第一段，sagin的优点、组成
% 参考了sagin-ad
SAGIN brings together aerial, satellite, and terrestrial communication networks to create a seamless, multi-layered system. This integration offers secure and dependable access to high-speed communication, making global connectivity available at all times and locations. SAGIN’s versatility enables it to support applications ranging from multimedia services and transportation to military operations, environmental monitoring, and emergency disaster response, ensuring robust communication for critical needs worldwide. As shown in Figure \ref{fig:sagin}, SAGIN is a multi-layered, hierarchical network consisting of three primary layers: the ground layer, air layer, and space layer. In the ground layer, mobile user equipment, base stations, and various IoT devices collectively form a robust foundation for connectivity and extensive data collection. These IoT devices play a crucial role in gathering environmental, operational, and situational data. The air layer, represented by Autonomous Aerial Vehicles (AAVs), serves as a flexible and dynamic extension of the network, enhancing coverage and connectivity, especially in remote or mobile scenarios. At the top, the space layer comprises a network of satellites that provides broad coverage.

% add fig 1

\subsection{Federated Learning Model}
\label{flmodel}

% ----------------- dfl的建模 -----------------
% 节点集合。异构性。
% ----- 关于空基和地基agent的描述 -----
As shown in Figure \ref{fig:sagin}, we consider a decentralized federated learning system over SAGIN mainly consisting of a set $\mathcal{A}$ of $N$ heterogeneous agents. The system accounts for both the dynamic evolution of the network topology and the inherent heterogeneity of the agent. Each participating agent, representing an air node (such as an AAV), is characterized by distinct capabilities that may vary in factors such as processing power, communication bandwidth, energy constraints, etc. Specifically, each agent $a_i \in \mathcal{A}, i \in N$ undertakes three primary roles:
% or a ground node (such as an Internet of Things (IoT) device). 
\begin{itemize}
    \item Local Data Collection: Each AAV agent monitors a designated area, overseeing the ground-based IoT devices within its range. These IoT devices collect critical data, such as images of disaster-affected regions for assessing infrastructure damage. Let each AAV agent $a_i$ holds a local dataset $D_i$, where the data size at each agent is denoted by $|D_i|$. 

    \item Local Model Training: The objective of each agent’s training process is to produce a local model update exclusively based on local data, without sharing raw data with other agents. Through the iterative exchange of model parameters with neighboring agents, each agent works towards establishing a local consensus, gradually aligning model representations across the network.

    \item Asynchronous Communication: After local training, each agent $a_i$ asynchronously communicates with its neighboring agents in the $t^{th}$ iteration, $\mathcal{N}_i^t$, which are determined by the dynamically evolving network topology represented by $\mathbf{W}^t$. In each iteration, only a subset of active agents $\mathcal{Q}^t$ is required to perform the computationally intensive gradient descent, allowing the process to proceed without waiting for all agents. Specifically, each agent $i$ is active with probability $p_i^t$, where its activation probability is related to its own capability. This asynchronous communication framework ensures that the global training speed is not hindered by slower agents.
\end{itemize}

It is noteworthy that the role/task assigned to each agent undergoes iterative changes over time. Moreover, at any given point, the role being executed by each agent is not homogeneous; instead, it is determined independently based on the agent's local state and available resources.

\subsection{Problem Formulation}
\label{preliminary}
Let $\mathbf{x}_i \in \mathbb{R}^n$ and $\mathbf{y}_i \in \mathbb{R}^m$ denote the upper-level and lower-level optimization variables of agent $i$, this paper considers the following decentralized federated bilevel optimization problem:
% \vspace{-0.5em}
\begin{equation}
\setlength{\abovedisplayskip}{2pt}
\setlength{\belowdisplayskip}{2pt}
% \vspace{-1em}
\begin{aligned}
\underset{\mathbf{x}}{\min} & \  \underset{i=1}{\overset{N}{\sum}} F_i(\mathbf{x}_i, \mathbf{y}_i) \triangleq G_i(\mathbf{x}_i,\mathbf{y}_i) + R(\mathbf{x}_i)\\ 
\text{s.t.} &\ \mathbf{x}_i=\mathbf{x}_j,  \forall i \in [N], \forall j \in \mathcal{N}_i\\
& \text{ }  \mathbf{y} = \underset{\mathbf{y}'}{\arg\min} \text{ } \underset{i=1}{\overset{N}{\sum}} f_i(\mathbf{x}_i, \mathbf{y}_i') \triangleq g_i(\mathbf{x}_i,\mathbf{y}_i') + r(\mathbf{y}_i')\\
& \text{ }  \text{s.t.} \ \mathbf{y}_i'=\mathbf{y}_j', \forall i \in [N], \forall j \in \mathcal{N}_i,\\
\end{aligned}
\label{eq1}
\end{equation}
where $\mathbf{x} \triangleq [\mathbf{x}_1, \cdots, \mathbf{x}_N]^{\top}$ and $\mathbf{y} \triangleq [\mathbf{y}_1, \cdots, \mathbf{y}_N]^{\top}$. $N$ is the number of agents on a connected graph $(\mathcal{V}, \mathcal{E})$, where $\mathcal{V}$ denotes the set of vertices (agents) and $\mathcal{E}$ denotes the set of edges (communication between agents). The graph can be represented by a mixing matrix $\mathbf{W} \in \mathbb{R}^{N \times N}$. $G_i(\cdot)$ and $g_i(\cdot)$ are smooth, possibly non-convex functions known only to agent $i$; $R(\cdot)$ and $r(\cdot)$ are convex, possibly non-smooth functions common to all agents. $F_i(\cdot)$ and $f_i(\cdot)$ denote the local objective functions. In practical computation, $G_i(\cdot)$ and $g_i(\cdot)$ take the form $G_i(\cdot;D_i)$ and $g_i(\cdot;D_i)$, indicating that they are specifically computed using the data available locally to each agent. $\mathcal{N}_i$ denotes the set of neighbors of agent $i$. The equality constraints $\mathbf{x}_i=\mathbf{x}_j$ and $\mathbf{y}_i'=\mathbf{y}_j'$, $\forall i \in [N]$, $\forall j \in \mathcal{N}_i$, enforce the model agreements (``consensus'') \cite{lu2022stochastic}.

\section{Proposed Algorithm: Argus}
\label{Argus}
In the proposed DFL framework over SAGIN, each AAV agent oversees a specific region of ground-level IoT devices, collecting data to train local models. These agents collaboratively address machine learning tasks, particularly bilevel optimization problems, which are valuable in SAGIN. For example, bilevel optimization techniques like meta-learning and hyperparameter tuning enhance adaptability and flexibility: meta-learning improves response to data heterogeneity across agents, while hyperparameter tuning enables swift model adjustments, critical for effective decision-making in remote disaster relief operations.

% 介绍整体的求解思路
To efficiently address the NP-hard bilevel optimization problem, Argus algorithm reformulates the original bilevel optimization into a single-level problem by embedding the lower-level problem as a constraint within the upper-level problem. Argus then constructs a set of cutting planes, a type of linear constraint, to approximate the feasible region of the upper-level problem. Specifically, agents iteratively aggregate information from their neighbors, execute local updates asynchronously, and periodically update the cutting planes. The detailed process is presented below.

\subsection{Lower-Level Estimation}
For the problem in Eq.(\ref{eq1}), we define the optimal solution of the lower-level problem as $\mathbf{y}^* = \arg \min_{\mathbf{y}'}\{\sum_{i=1}^N f_i(\mathbf{x}_i,\mathbf{y}_i') :\mathbf{y}_i'=\mathbf{y}_j', \forall i \in [N], \forall j \in \mathcal{N}_i\}$ and $h(\mathbf{x}, \mathbf{y}) = \left \| \mathbf{y} - \mathbf{y}^* \right \|_1 + \lambda_1\left \| \mathbf{y} - \mathbf{y}^* \right \|_2^2$, where $\lambda_1$ is a weight coefficient. Then we can reformulate the problem in Eq.(\ref{eq1}) as the following single-level optimization problem with constraints:
%\vspace{-0.8em}
%\begin{small}
\begin{equation}
% \vspace{-0.8em}
\setlength{\abovedisplayskip}{2pt}
\setlength{\belowdisplayskip}{2pt}
\begin{aligned}
\underset{
\mathbf{x}, \mathbf{y}}{\min} &  \underset{i=1}{\overset{N}{\sum}} F_i(\mathbf{x}_i, \mathbf{y}_i) \triangleq G_i(\mathbf{x}_i,\mathbf{y}_i) + R(\mathbf{x}_i)\\
\text{s.t.} &  \text{ }\mathbf{x}_i=\mathbf{x}_j,  \forall i \in [N], \forall j \in \mathcal{N}_i \\
& \text{ }  h(\mathbf{x}, \mathbf{y}) = 0.
\end{aligned}
\label{eq4}
\end{equation}
%\end{small}

Focusing on $\mathbf{y}^{*}$ in the constraint $h(\mathbf{x}, \mathbf{y}) = 0$, and given the inherent difficulty in directly solving the non-convex lower-level problem, we approximate $\mathbf{y}^{*}$ as follows. Firstly, we obtain the first-order Taylor approximation of $g_{i}(\mathbf{x}_i,\mathbf{y}_{i}')$ with respect to $\mathbf{x}_i$. That is, for a given point $\bar{\mathbf{x}}_i$, $\tilde{g}_{i}(\mathbf{x}_i,\mathbf{y}_{i}') = g_{i}(\bar{\mathbf{x}}_i,\mathbf{y}_{i}') + \nabla_{\mathbf{x}_i}g_{i}(\bar{\mathbf{x}}_i,\mathbf{y}_{i}')^{\top}(\mathbf{x}_i-\bar{\mathbf{x}}_i)$, $\tilde{f}_i(\mathbf{x}_i,\mathbf{y}_i')=\tilde{g}_{i}(\mathbf{x}_i,\mathbf{y}_{i}')+r(\mathbf{y}_i')$. Then the augmented Lagrangian function of the lower-level optimization problem in Eq.(\ref{eq1}) is 
%\vspace{-0.8em}
%\begin{small}
\begin{equation}
% \vspace{-0.5em}
% \setlength{\abovedisplayskip}{2pt}
% \setlength{\belowdisplayskip}{2pt}
\begin{aligned}
g_{p}(\mathbf{x}, \mathbf{y}', \boldsymbol{\varphi})  & = \sum_{i=1}^{N}
 \tilde{f}_i(\mathbf{x}_i,\mathbf{y}_i')\\ & +
 \sum_{i=1}^{N}\sum_{j\in \mathcal{N}_i} (
 \boldsymbol{\varphi}_{ij}^\top (\mathbf{y}_{i}'-\mathbf{y}_{j}') + \frac{\mu}{2}||\mathbf{y}_{i}'-\mathbf{y}_{j}'||^2_2 ), 
\end{aligned}
\label{eq6.0}
\end{equation}
%\end{small}
where $\boldsymbol{\varphi}$ is the dual variable and $\boldsymbol{\varphi}_{ij} \in \mathbb{R}^m$. $\mu >0$ is a penalty parameter. Inspired by \cite{shi2015proximal}, to cope with the non-smooth term, we use a variant of Eq.(\ref{eq6.0}) as follows:
%\vspace{-0.5em}
%\begin{small}
\begin{equation}
%\vspace{-0.5em}
% \setlength{\abovedisplayskip}{2pt}
% \setlength{\belowdisplayskip}{2pt}
\begin{aligned}
g_{p}'(\mathbf{x}, \mathbf{y}', \boldsymbol{\varphi}) &=  \sum_{i=1}^{N}
 \tilde{g}_i(\mathbf{x}_i,\mathbf{y}_i')
 \\ &+\sum_{i=1}^{N}\sum_{j\in \mathcal{N}_i} (
 \boldsymbol{\varphi}_{ij}^\top (\mathbf{y}_{i}'-\mathbf{y}_{j}') + \frac{\mu}{2}||\mathbf{y}_{i}'-\mathbf{y}_{j}'||^2_2 ).
\end{aligned}
\label{eq6}
\end{equation}
%\end{small}

Then we can use $K$ communication rounds to approximate $\mathbf{y}^*$. Specifically, in the $(k+1)^{th}$ round, each agent communicates with neighbors and updates variables as follows:
% 分开各节点写的版本
%\vspace{-0.5em}
\begin{equation}
%\vspace{-0.5em}
\begin{aligned}
{\mathbf{y}'_{i}}^{(k+1)} & = \text{prox}_{r}(\sum_{j\in \mathcal{N}_i}\mathbf{W}_{ij}{\mathbf{y}'}_{j}^{(k)} \\
& - \eta_{y} \nabla_{\mathbf{y}_i} g_{p}'(\mathbf{x}, {\mathbf{y}'}^{(k)}, \boldsymbol{\varphi}^{(k)})),
\end{aligned}
\label{eq7}
\end{equation}

% \vspace{-0.5em}
\begin{equation}
%\vspace{-0.5em}
\begin{aligned}
\boldsymbol{\varphi}_{ij}^{(k+1)} =\boldsymbol{\varphi}_{ij}^{(k)} + \eta_{\boldsymbol{\varphi}}\nabla_{\boldsymbol{\varphi}_{ij}} g_{p}'(\mathbf{x},{\mathbf{y}'}^{(k+1)},\boldsymbol{\varphi}^{(k)}),
\end{aligned}
\label{eq8}
\end{equation}
% u-v前面没有eta，r改成eta和N
where $\eta_y$ and $\eta_{\boldsymbol{\varphi}}$ are step-sizes. $\text{prox}_r = {\arg\min}_{\mathbf{u}} \{r(\mathbf{u})+\frac{1}{2}\left\|\mathbf{u}-\mathbf{v}\right\|^{2}\}$ denotes the proximal operator. Then each agent broadcasts ${\mathbf{y}'_{i}}^{(k+1)}$ and $\boldsymbol{\varphi}_{ij}^{(k+1)}$ to neighbors.

The results after $K$ rounds are utilized to approximate $\mathbf{y}^*$: 
%\vspace{-0.8em}
\begin{equation}
%\vspace{-0.8em}
% \setlength{\abovedisplayskip}{2pt}
% \setlength{\belowdisplayskip}{2pt}
    \mathbf{y}^* = {\mathbf{y}'}^{(K)}.
    \label{eq10}
\end{equation}

\subsection{Polyhedral Approximation}
\label{Polyhedral}
Based on the approximated solution of the lower-level problem, we can obtain a relaxed problem with respect to Eq.(\ref{eq4}) as follows:

%\vspace{-1.5em}
%\begin{small}
\begin{equation}
%\vspace{-0.5em}
% \setlength{\abovedisplayskip}{2pt}
% \setlength{\belowdisplayskip}{2pt}
\begin{aligned}
\underset{
\mathbf{x}, \mathbf{y}}{\min} & \text{ }\underset{i=1}{\overset{N}{\sum}} F_i(\mathbf{x}_i, \mathbf{y}_i) \triangleq G_i(\mathbf{x}_i,\mathbf{y}_i) + R(\mathbf{x}_i)\\
\text{s.t.} & \text{ }\mathbf{x}_i=\mathbf{x}_j,  \forall i \in [N], \forall j \in \mathcal{N}_i \\
&  \text{ } h(\mathbf{x}, \mathbf{y}) \le \varepsilon,
\end{aligned}
\label{eq11.5}
\end{equation}
%\end{small}
% 凸性分析
where $\varepsilon >0$ is a constant. By setting $K=1$ in Eq.(\ref{eq10}) and according to the operations that preserve convexity \cite{boyd2004convex}, $h(\mathbf{x}, \mathbf{y})$ is a convex function. Then the feasible region with respect to constraint $h(\mathbf{x}, \mathbf{y}) \leq \varepsilon$ in Eq.(\ref{eq11.5}) is a convex set \cite{jiao2022asynchronous}, and we can utilize the cutting plane method to approximate it. The cutting plane method iteratively approximates the feasible region of an optimization problem by introducing cutting planes that exclude infeasible solutions from the approximated polyhedron. Specifically, each agent maintains a set of cutting planes that can form a polytope $\boldsymbol{\mathcal{P}}$. The polytope of agent $i$ in the $(t + 1)^{th}$ iteration is:
%\begin{small}
%\vspace{-0.5em}
\begin{equation}
%\vspace{-0.2em}
% \setlength{\abovedisplayskip}{2pt}
% \setlength{\belowdisplayskip}{2pt}
\boldsymbol{{\mathcal{P}}}_i^t = \{ \sum_{j\in \mathcal{N}_i} \boldsymbol{a}_{j,l}^{\top} \mathbf{x}_j + \sum_{j\in \mathcal{N}_i}\boldsymbol{b}_{j,l}^{\top} \mathbf{y}_j + c_{l} \leq 0, \forall l \in [|\boldsymbol{{\mathcal{P}}}_i^t|] \},
\label{eq12}
\end{equation}
%\end{small}
where $\{\boldsymbol{a}_{j,l} \in \mathbb{R}^n\}, \{\boldsymbol{b}_{j,l} \in \mathbb{R}^m\}$ and $c_{l} \in \mathbb{R}^1$ denote the parameters of the $l^{th}$ cutting plane in agent $i$. These parameters are calculated according to the definitions of a valid cutting plane and the convex properties of $h(\mathbf{x}, \mathbf{y})$. More details are introduced in Section \ref{Argus}.C.(3). $|\boldsymbol{{\mathcal{P}}}_i^t|$ is the number of cutting planes. We set that $|\boldsymbol{{\mathcal{P}}}_i^t|<M, \forall t$. Then we have an approximate problem as:
%\vspace{-0.8em}
%\begin{small}
\begin{equation}
\vspace{-0.2em}
\begin{aligned}
\underset{
\mathbf{x}, \mathbf{y}}{\min} & \text{ } \underset{i=1}{\overset{N}{\sum}} F_i(\mathbf{x}_i, \mathbf{y}_i) \triangleq G_i(\mathbf{x}_i,\mathbf{y}_i) + R(\mathbf{x}_i)\\
\text{s.t.} & \text{ } \mathbf{x}_i=\mathbf{x}_j, \forall i \in [N], \forall j \in \mathcal{N}_i \\
& \text{ } \underset{j\in \mathcal{N}_i}{\overset{}{\sum}} \boldsymbol{a}_{j,l}^{\top} \mathbf{x}_j + \underset{j\in \mathcal{N}_i}{\overset{}{\sum}}  \boldsymbol{b}_{j,l}^{\top} \mathbf{y}_j + c_{l}  \leq 0, \forall i \in [N], \forall l \in [|\boldsymbol{{\mathcal{P}}}_i^t|].
\end{aligned}
\label{eq14}
\end{equation}
%\end{small}

\subsection{Asynchronous Decentralized Updating}
To solve the constrained optimization problem with a non-convex objective function in Eq.(\ref{eq14}), we first formulate its Lagrangian function:
%\vspace{-0.8em}
\begin{equation}
%\vspace{-0.8em}
% \setlength{\abovedisplayskip}{2pt}
% \setlength{\belowdisplayskip}{2pt}
\begin{aligned}
L_{p}(\mathbf{x}, \mathbf{y}, \lambda, \boldsymbol{\theta}) &= \underset{i=1}{\overset{N}{\sum}} F_i(\mathbf{x}_i,\mathbf{y}_i) 
+  \underset{i=1}{\overset{N}{\sum}} \underset{l=1}{\overset{|\boldsymbol{{\mathcal{P}}}_i^t|}{\sum}} \lambda_{i,l} (\underset{j\in \mathcal{N}_i}{\overset{}{\sum}} \boldsymbol{a}_{j,l}^{\top} \mathbf{x}_j \\
&+ \underset{j\in \mathcal{N}_i}{\overset{}{\sum}}  \boldsymbol{b}_{j,l}^{\top} \mathbf{y}_j + c_{l} ) + \underset{i=1}{\overset{N}{\sum}}\underset{j\in \mathcal{N}_i}{\overset{}{\sum}} \boldsymbol{\theta}_{i,j}^\top(\mathbf{x}_i-\mathbf{x}_j),
\end{aligned}
\label{eq15}
\end{equation}
where $\lambda$ and $\boldsymbol{\theta}$ are dual variables, $\lambda_{i,l} \in \mathbb{R}^1$, $\boldsymbol{\theta}_{i,j} \in \mathbb{R}^n$. The presence of $R(\cdot)$ may render $F_i(\cdot)$ non-differentiable, so we substitute $F_i(\cdot)$ in Eq.(\ref{eq15}) with $G_i(\cdot)$ to obtain:
%\begin{small}
%\vspace{-0.8em}
\begin{equation}
%\vspace{-0.8em}
% \setlength{\abovedisplayskip}{2pt}
% \setlength{\belowdisplayskip}{2pt}
\begin{aligned}
L_{p}'(\mathbf{x}, \mathbf{y}, \lambda, \boldsymbol{\theta}) &= \underset{i=1}{\overset{N}{\sum}} G_i(\mathbf{x}_i,\mathbf{y}_i) 
+  \underset{i=1}{\overset{N}{\sum}} \underset{l=1}{\overset{|\boldsymbol{{\mathcal{P}}}_i^t|}{\sum}} \lambda_{i,l} (\underset{j\in \mathcal{N}_i}{\overset{}{\sum}} \boldsymbol{a}_{j,l}^{\top} \mathbf{x}_j \\ & + \underset{j\in \mathcal{N}_i}{\overset{}{\sum}}  \boldsymbol{b}_{j,l}^{\top} \mathbf{y}_j + c_{l} )+ \underset{i=1}{\overset{N}{\sum}}\underset{j\in \mathcal{N}_i}{\overset{}{\sum}} \boldsymbol{\theta}_{i,j}^\top(\mathbf{x}_i-\mathbf{x}_j).
\end{aligned}
\label{eq15.1}
\end{equation}
%\end{small}

Then, following \cite{xu2023unified}, a regularized version of Eq.(\ref{eq15.1}) can be expressed as: 
%\vspace{-1em}
%\begin{small}
\begin{equation}
\setlength{\abovedisplayskip}{2pt}
\setlength{\belowdisplayskip}{2pt}
\begin{aligned}
&\tilde{L}_p(\mathbf{x}, \mathbf{y}, \lambda, \boldsymbol{\theta}) \\
& = L_{p}'(\mathbf{x}, \mathbf{y}, \lambda, \boldsymbol{\theta}) - \sum_{i=1}^N\sum_{l=1}^{|\boldsymbol{{\mathcal{P}}}_i^t|}\frac{c_1^t}{2}||\lambda_{i,l}||^2  - \sum_{i=1}^N \underset{j\in \mathcal{N}_i}{\overset{}{\sum}}\frac{c_2^t}{2}||\boldsymbol{\theta}_{i,j}||^2 \\ 
& =\sum_{i=1}^{N} \tilde{L}_{pi}(\{\mathbf{x}_i\}, \{\mathbf{y}_i\},\left \{\lambda_{i,l} \right \},\{\boldsymbol{\theta}_{i,j}\}),
\end{aligned}
\label{eq17}
\end{equation}
%\end{small}

where 
\begin{equation}
\setlength{\abovedisplayskip}{2pt}
\setlength{\belowdisplayskip}{2pt}
\begin{aligned}
& \tilde{L}_{pi}(\{\mathbf{x}_i\}, \{\mathbf{y}_i\},\left \{\lambda_{i,l} \right \},\{\boldsymbol{\theta}_{i,j}\}) \\
& = G_i(\mathbf{x}_i,\mathbf{y}_i) 
+  \underset{l=1}{\overset{|\boldsymbol{{\mathcal{P}}}_i^t|}{\sum}} \lambda_{i,l} (\underset{j\in \mathcal{N}_i}{\overset{}{\sum}} \boldsymbol{a}_{j,l}^{\top} \mathbf{x}_j + \underset{j\in \mathcal{N}_i}{\overset{}{\sum}}  \boldsymbol{b}_{j,l}^{\top} \mathbf{y}_j + c_{l} ) \\ &+ \underset{j\in \mathcal{N}_i}{\overset{}{\sum}} \boldsymbol{\theta}_{i,j}^\top(\mathbf{x}_i-\mathbf{x}_j) - \sum_{l=1}^{|\boldsymbol{{\mathcal{P}}}_i^t|}\frac{c_1^t}{2}||\lambda_{i,l}||^2 - \underset{j\in \mathcal{N}_i}{\overset{}{\sum}}\frac{c_2^t}{2}||\boldsymbol{\theta}_{i,j}||^2,
\end{aligned}
\label{eq17+1}
\end{equation}
and $c_1^t=\frac{1}{\eta_{\lambda}(t+1)^{\frac{1}{2}}}$ and $c_2^t=\frac{1}{\eta_{\theta}(t+1)^{\frac{1}{2}}}$ are non-negative non-increasing regularization sequences.

% while slow agents only perform simple aggregation operations
To deal with the challenge of stragglers, Argus updates variables in an asynchronous manner. In each iteration, only a set of fast agents $\mathcal{Q}$ complete computationally intensive gradient calculations. The proposed algorithm proceeds as follows.

\noindent \textbf{(1) Consensus update and (proximal) gradient descent}. Agents first communicate with neighbors and aggregate information as follows,
%\vspace{-0.5em}
\begin{equation}
\setlength{\abovedisplayskip}{2pt}
\setlength{\belowdisplayskip}{2pt}
\mathbf{d}_i^{t} = \sum_{j\in \mathcal{N}_i}\mathbf{W}_{ij}^{t+1}\mathbf{x}_j^{t},
\label{asyeq_1}
\end{equation}
%\vspace{-0.5em}
\begin{equation}
\setlength{\abovedisplayskip}{2pt}
\setlength{\belowdisplayskip}{2pt}
\mathbf{u}_i^{t} = \sum_{j\in \mathcal{N}_i}\mathbf{W}^{t+1}_{ij}\mathbf{y}_{j}^{t}.
\label{asyeq_2}
\end{equation}

Then the local update rule is:
%\hspace{5em} 
%\vspace{-0.5em}
\begin{equation}
\setlength{\abovedisplayskip}{2pt}
\setlength{\belowdisplayskip}{2pt}
\mathbf{x}_{i}^{t+1} = \left\{\begin{aligned}
&\underset{\mathbf{x}_i}{\text{argmin}}\{\frac{1}{2}||\mathbf{x}_i - (\mathbf{d}_{i}^{t}-\eta^t_{i,x} \nabla_{\mathbf{x}_i} \tilde{L}_{pi}(\{\mathbf{x}_i^{\hat{t}_i}\}, \{\mathbf{y}_i^{\hat{t}_i}\}, \\
&\hspace{2.4em}  \{\lambda_{i,l}^{\hat{t}_i} \},\{\boldsymbol{\theta}^{\hat{t}_i}_{i,j}\}))||^2 + \eta^t_{i,x}R(\mathbf{x}_i)\}, i \in \mathcal{Q}^{t+1};\\
&\mathbf{d}_i^{t}, i \notin \mathcal{Q}^{t+1},
\end{aligned}\right.
\label{asyeq_3}
\end{equation}
%\vspace{-0.5em}
\begin{equation}
\setlength{\abovedisplayskip}{2pt}
\setlength{\belowdisplayskip}{2pt}
\mathbf{y}_{i}^{t+1} = \left\{\begin{aligned}
&\mathbf{u}_i^{t}-\eta^t_{i,y} \nabla_{\mathbf{y}_i} \tilde{L}_{pi}(\{\mathbf{x}_i^{\hat{t}_i}\}, \{\mathbf{y}_i^{\hat{t}_i}\}, \{\lambda_{i,l}^{\hat{t}_i} \},\{\boldsymbol{\theta}^{\hat{t}_i}_{i,j}\}),\\ &
\hspace{14.7em} i \in \mathcal{Q}^{t+1}; \\
&\mathbf{u}_i^{t}, i \notin \mathcal{Q}^{t+1},
\end{aligned}\right.
\label{asyeq_4}
\end{equation}
where $\eta^t_{i,x}$ and $\eta^t_{i,y}$ are learning rates \cite{jeong2022asynchronous}, $\hat{t}_i$ denotes the last iteration during which agent $i$ was active. Then, each agent broadcasts the local variables $\mathbf{x}_{i}^{t+1}$ and $\mathbf{y}_{i}^{t+1}$ to neighbors.

\noindent \textbf{(2) The update of dual variables}. For agents in $\mathcal{Q}^{t+1}$, they update dual variables as follows,
%\vspace{-0.5em}
%\begin{small}
\begin{equation}
\setlength{\abovedisplayskip}{2pt}
\setlength{\belowdisplayskip}{2pt}
{\lambda}_{i,l}^{t+1}={\lambda}_{i,l}^{t} + \eta_{i,\lambda}^t\nabla_{\lambda_{i,l}}\tilde{L}_{pi}(\{\mathbf{x}_i^{t+1}\}, \{\mathbf{y}_i^{t+1}\},\{ \boldsymbol{\lambda}_{i,l}^{\hat{t}_i} \},\{\boldsymbol{\theta}_{i,j}^{\hat{t}_i}\}),
\label{asyeq_5}
\end{equation}
%\end{small}
%\vspace{-0.5em}
%\begin{small}
\begin{equation}
\setlength{\abovedisplayskip}{2pt}
\setlength{\belowdisplayskip}{2pt}
\boldsymbol{\theta}_{i,j}^{t+1}=\boldsymbol{\theta}_{i,j}^{t} + \eta_{i,\theta}^t\nabla_{\boldsymbol{\theta}_{i,j}}\tilde{L}_{pi}(\{\mathbf{x}_i^{t+1}\}, \{\mathbf{y}_i^{t+1}\},\{\lambda_{i,l}^{t+1} \},\{\boldsymbol{\theta}_{i,j}^{\hat{t}_i}\}),
\label{asyeq_6}
\end{equation}
%\end{small}
where $\eta_{i,\lambda}^t$ and $\eta_{i,\theta}^t$ are step-sizes. Finally, each active agent broadcasts ${\lambda}_{i,l}^{t+1}$ and $\boldsymbol{\theta}_{i,j}^{t+1}$ to neighbors.

% in the case we consider, $h(\mathbf{x}, \mathbf{y})$ is a convex function. Therefore
\noindent \textbf{(3) The update of cutting planes.} Before the $T_1^{th}$ iteration, agents update cutting planes every $\iota$ iterations based on local and neighbor information. In agent $i$, given that $j \in \mathcal{N}_i$, if $(\{\mathbf{x}_j^{t+1}\}, \{\mathbf{y}_j^{t+1}\})$ is not feasible for $h(\{\mathbf{x}_j\}, \{\mathbf{y}_j\}) = \left \| \{\mathbf{y}_j\} - \{\mathbf{y}_j^*\} \right \|_1 + \lambda_1\left \| \{\mathbf{y}_j\} - \{\mathbf{y}_j^*\} \right \|_2^2 \leq \varepsilon$, we aim to find a cutting plane to separate $(\{\mathbf{x}_j^{t+1}\}, \{\mathbf{y}_j^{t+1}\})$ from the feasible region of the local constraint $h(\{\mathbf{x}_j\}, \{\mathbf{y}_j\}) \leq \varepsilon$. Specifically, a valid cutting plane \cite{franc2011cutting} $\sum_{j\in \mathcal{N}_i} \boldsymbol{a}_{j}^{\top} \mathbf{x}_j + \sum_{j\in \mathcal{N}_i}\boldsymbol{b}_{j}^{\top} \mathbf{y}_j + c \leq 0$ satisfies that
%\hspace{9.5em}
%\vspace{-0.5em}
\begin{small}
\begin{equation}
\setlength{\abovedisplayskip}{2pt}
\setlength{\belowdisplayskip}{2pt}
\left\{\begin{aligned}
& \sum\nolimits_{j\in \mathcal{N}_i} \boldsymbol{a}_{j}^{\top} \mathbf{x}_j + \sum\nolimits_{j\in \mathcal{N}_i}\boldsymbol{b}_{j}^{\top} \mathbf{y}_j + c\leq 0, \\ 
& \hspace{6em} \forall (\{\mathbf{x}_j\}, \{\mathbf{y}_j\}) \text{ }\text{satisfies}\text{ } h(\{\mathbf{x}_j\}, \{\mathbf{y}_j\}) \leq \varepsilon; \\
&\sum\nolimits_{j\in \mathcal{N}_i}\boldsymbol{a}_{j}^{\top} \mathbf{x}_j + \sum\nolimits_{j\in \mathcal{N}_i}\boldsymbol{b}_{j}^{\top} \mathbf{y}_j + c > 0.
\end{aligned}\right.
\label{eq25}
\end{equation}
\end{small}
As discussed in Section \ref{Polyhedral}, according to the properties of $h(\cdot)$, we have:
%\vspace{-0.5em}
%\begin{footnotesize}
\begin{equation}
\begin{aligned}
\setlength{\abovedisplayskip}{2pt}
\setlength{\belowdisplayskip}{2pt}
&h(\{\mathbf{x}_j\}, \{\mathbf{y}_j\}) \ge h(\{\mathbf{x}_j^{t+1}\}, \{\mathbf{y}_j^{t+1}\})  \\
&\hspace{2em} + \left[\begin{aligned}
\{\frac{\partial h(\{\mathbf{x}_j^{t+1}\}, \{\mathbf{y}_j^{t+1}\})}{\partial \mathbf{x}_j}\} \\
\{\frac{\partial h(\{\mathbf{x}_j^{t+1}\}, \{\mathbf{y}_j^{t+1}\})}{\partial \mathbf{y}_j} \}
\end{aligned}\right]^{\top} 
\left(\left[\begin{aligned}
\{\mathbf{x}_j\} \\
\{\mathbf{y}_j\} \\
\end{aligned}\right]-
\left[\begin{aligned}
\{\mathbf{x}_j^{t+1}\} \\
\{\mathbf{y}_j^{t+1}\} \\
\end{aligned}\right]\right).
\label{eq26}
\end{aligned}
\end{equation}
%\end{footnotesize}

Based on Eq.(\ref{eq25}) with Eq.(\ref{eq26}), we can find a valid cutting plane $cp_{new}^{t+1}$ w.r.t. $(\{\mathbf{x}_j^{t+1}\}, \{\mathbf{y}_j^{t+1}\})$ as 
%\vspace{-0.5em}
%\begin{footnotesize}
\begin{equation}
\begin{aligned}
\setlength{\abovedisplayskip}{2pt}
\setlength{\belowdisplayskip}{2pt}
& h(\{\mathbf{x}_j^{t+1}\}, \{\mathbf{y}_j^{t+1}\}) \\
& + \left[\begin{aligned}
\{\frac{\partial h(\{\mathbf{x}_j^{t+1}\}, \{\mathbf{y}_j^{t+1}\})}{\partial \mathbf{x}_j}\} \\
\{\frac{\partial h(\{\mathbf{x}_j^{t+1}\}, \{\mathbf{y}_j^{t+1}\})}{\partial \mathbf{y}_j} \}
\end{aligned}\right]^{\top} 
\left(\left[\begin{aligned}
\{\mathbf{x}_j\} \\
\{\mathbf{y}_j\} \\
\end{aligned}\right]-
\left[\begin{aligned}
\{\mathbf{x}_j^{t+1}\} \\
\{\mathbf{y}_j^{t+1}\} \\
\end{aligned}\right]\right)
\le \varepsilon.
\end{aligned}
\label{eq27}
\end{equation}
%\end{footnotesize}

In other words, the parameters of the new cutting plane are:
\begin{equation}
\begin{aligned}
\mathbf{a}_j & = \frac{\partial h(\{\mathbf{x}_j^{t+1}\}, \{\mathbf{y}_j^{t+1}\})}{\partial \mathbf{x}_j},
\end{aligned}
\label{eqcp_a}
\end{equation}

\begin{equation}
\begin{aligned}
\mathbf{b}_j & = \frac{\partial h(\{\mathbf{x}_j^{t+1}\}, \{\mathbf{y}_j^{t+1}\})}{\partial \mathbf{y}_j},
\end{aligned}
\label{eqcp_b}
\end{equation}

\begin{equation}
\begin{aligned}
c & =  h(\{\mathbf{x}_j^{t+1}\}, \{\mathbf{y}_j^{t+1}\}) \\
&-  \left[\begin{aligned}
\{\frac{\partial h(\{\mathbf{x}_j^{t+1}\}, \{\mathbf{y}_j^{t+1}\})}{\partial \mathbf{x}_j}\} \\
\{\frac{\partial h(\{\mathbf{x}_j^{t+1}\}, \{\mathbf{y}_j^{t+1}\})}{\partial \mathbf{y}_j} \}
\end{aligned}\right]^{\top} 
\left(\left[\begin{aligned}
\{\mathbf{x}_j^{t+1}\} \\
\{\mathbf{y}_j^{t+1}\} \\
\end{aligned}\right]\right)
- \varepsilon.
\end{aligned}
\label{eqcp_c}
\end{equation}

That is, if $h (\{\mathbf{x}_j^{t+1}\}, \{\mathbf{y}_j^{t+1}\}) >\varepsilon$, $cp_{new}^{t+1}$ is added to $\boldsymbol{{\mathcal{P}}}_i^{t+1}$, and $\lambda_{|\boldsymbol{{\mathcal{P}}}_i^{t+1}|}^{t+1}$ is added to $\{\lambda_i^{t+1}\}$. Besides, when $\lambda_{i,l}^{t+1}=0$ and $\lambda_{i,l}^{t}=0$, the $l^{th}$ cutting plane is considered inactive, leading to $cp_{l}$ and $\lambda_{i,l}$ be removed. 
Finally, agents broadcast $\boldsymbol{{\mathcal{P}}}_i^{t+1}$ and $\{\lambda_{i}^{t+1}\}$ to neighbors. The details of the proposed algorithm are summarized in Algorithm \ref{algo}.

\begin{algorithm} 
\caption{Argus} 
\label{algo}
\begin{algorithmic}
\STATE \textbf{Initialization:} local variables $\mathbf{x}^0$, $\mathbf{y}^0$, $\lambda^0$, $\boldsymbol{\theta}^0$, $\mathbf{d}^0$, $\mathbf{u}^0$; polytope $\{\boldsymbol{{\mathcal{P}}}_i^{0}\}$.
\REPEAT
\STATE Agents communicate and update local variables according to Eq.(\ref{asyeq_3}) and Eq.(\ref{asyeq_4});
%\STATE Agents broadcast $\mathbf{x}_{i}^{t+1}$ and $\mathbf{y}_{i}^{t+1}$ to neighbors. 
\STATE Agents communicate and update dual variables according to Eq.(\ref{asyeq_5}) and Eq.(\ref{asyeq_6});
%\STATE Agents broadcast $\{{\lambda}_{i,l}^{t+1}\}$ and $\{\boldsymbol{\theta}_{i,j}^{t+1}\}$ to neighbors.  
\IF{$(t+1)$ mod $\iota == 0$ and $t<T_1$}
\STATE Agents compute $\mathbf{y}^*$ by Eq.(\ref{eq10});
\STATE Agents update $\boldsymbol{{\mathcal{P}}}_i^{t+1}$ and $\{\lambda_i^{t+1}\}$, $\forall i \in [N]$.
\ENDIF
\STATE $t = t + 1$;
\UNTIL{termination.}   
\end{algorithmic} 
\end{algorithm}

% 补充一个sagin的关联分析
% 描述一下上述算法步骤和sagin的关系
In summary, Argus is mainly performed by AAV agents. During each training iteration, active agents communicate with neighboring agents to aggregate parameters, update local models, and broadcast the new parameters. Upon receiving these updates, agents adjust dual variables and refine cutting planes every \(\iota\) iterations, repeating this cycle until termination. In the testing phase, each AAV agent utilizes its trained local model to serve its designated area.

% Additionally, dynamic links are established between a virtual satellite node and agents, enabling communication with neighbors at least once every $\tau$ rounds. This is critical for the convergence of the algorithm. 

\section{convergence analysis}
\label{discussion}

\emph{Assumption 1.} Following \cite{qian2019robust,ji2021bilevel, jiao2022asynchronous,mancino2022proximal, jeong2022asynchronous, lian2018asynchronous}, we assume that functions and variables satisfy:

(a) Lipschitzian gradient: Given that $L'_{pi}(\{\mathbf{x}_i\}, \{\mathbf{y}_i\},$ 
$\left \{\lambda_{i,l} \right \},\{\boldsymbol{\theta}_{i,j}\}) = G_i(\mathbf{x}_i,\mathbf{y}_i) +  \underset{l=1}{\overset{|\boldsymbol{{\mathcal{P}}}_i^t|}{\sum}} \lambda_{i,l} (\underset{j\in \mathcal{N}_i}{\overset{}{\sum}} \boldsymbol{a}_{j,l}^{\top} \mathbf{x}_j + \underset{j\in \mathcal{N}_i}{\overset{}{\sum}}  \boldsymbol{b}_{j,l}^{\top} \mathbf{y}_j$ 
$ + c_{l} ) + \underset{j\in \mathcal{N}_i}{\overset{}{\sum}} \boldsymbol{\theta}_{i,j}^\top(\mathbf{x}_i-\mathbf{x}_j)$, $\forall i \in [N]$, $L_{pi}'$ has Lipschitz continuous gradients, i.e., for any $\mathbf{a}, \mathbf{b}$, there exists $L > 0$ satisfying that $||\nabla L_{pi}'(\mathbf{a})-\nabla L_{pi}'(\mathbf{b})|| \le L||\mathbf{a}-\mathbf{b}||$.

(b) Convex proximal operator: $R$ and $r$ are convex, possibly non-smooth functions, such as $l_1$ norm. They admit proximal mappings that are easily computable.

(c) Boundedness: Dual variables are bounded, i.e., $||\lambda_{i,l}||^2 \le \alpha_1$, $||\boldsymbol{\theta}_{i,j}||^2 \le \alpha_2$. Before obtaining the $\epsilon$-stationary point, local variables satisfy that $\sum_{i=1}^N||\mathbf{x}_i^{t+1} - \mathbf{x}_i^t||^2 + \sum_{i=1}^N||\mathbf{y}_i^{t+1} - \mathbf{y}_i^t||^2 \ge \vartheta$, where $\vartheta>0$ is a relative small constant. Furthermore, we assume that each agent is capable of performing local gradient calculations and will be active at least once every $\tau$ iterations, which is reasonable given that agent heterogeneity in SAGIN is limited. Consequently, the change of the local variables is upper bounded within $\tau$ iterations: $\sum_{i=1}^N||\mathbf{x}_i^t - \mathbf{x}_i^{t-k}||^2\le\tau k_1 \vartheta$, $\sum_{i=1}^N||\mathbf{y}_i^t - \mathbf{y}_i^{t-k}||^2\le\tau k_1 \vartheta, \forall 1 \le k \le \tau$, where $k_1 > 0$ is a constant.

\emph{Assumption 2.} Following \cite{lian2018asynchronous, kovalev2021adom}, we assume the mixing matrix $\mathbf{W}^t$ satisfies the following properties:

(a) $\mathbf{W}^t_{ij}>0$ if $(i,j)\in \mathcal{E}^t$; otherwise $\mathbf{W}^t_{ij}=0$.

(b) $\mathbf{W}^t={\mathbf{W}^t}^\top$.

(c) null $(\mathbf{I}-\mathbf{W}^t)$ = span$\{\mathbf{e}\}$, where $\mathbf{e} \in \mathbb{R}^N$ is the vector of all ones.

(d) The eigenvalues of $\mathbf{W}^t$ lie in the range $(-1,1]$ with $\rho \triangleq\left\|\mathbf{W}^t-\frac{1}{N} \mathbf{e e}^{\top}\right\|_{2}<1$, where the value $\rho$ indicates the connectedness of the
graph \cite{mancino2022proximal}. 

We emphasize that Assumptions 2(d) pertains to the connectivity among agents, ensuring that agents can share information and reach consensus through the mixing matrix. Concerns may arise if an AAV becomes isolated due to distance or challenging terrain, potentially disrupting network connectivity. However, in SAGIN, the satellite network can provide relay services to reconnect isolated agents with their neighbors. Following the virtual node strategy in \cite{xiaogang2016survey}, each logical satellite position is consistently occupied by the nearest satellite, enabling agents to obtain connections anytime in necessity. Furthermore, with Low Earth Orbit (LEO) satellites positioned at altitudes of 500 to 2000 kilometers, the propagation delay to both airborne and terrestrial nodes is reduced to just a few milliseconds, thereby alleviating latency concerns in satellite communications \cite{wang2024resource}.

\emph{Definition 1.} (Proximal gradient mapping) Following standard definitions \cite{alghunaim2022unified}, given $\mathbf{a} \in $dom$(r)$, $\mathbf{b}$, and $\eta>0$, define the proximal gradient mapping of $\mathbf{b}$ at $\mathbf{a}$ to be
%\vspace{-0.5em}
%\begin{small}
\begin{equation}
\setlength{\abovedisplayskip}{2pt}
\setlength{\belowdisplayskip}{2pt}
P(\mathbf{a},\mathbf{b},\eta) \triangleq \frac{1}{\eta}(\mathbf{a}-\text{prox}_r (\mathbf{a}-\eta \mathbf{b})).
\label{eq30}
\end{equation}

% 待修改：需要引入虚拟学习率，并在收敛矩阵中使用
\emph{Definition 2.} (Virtual learning rate) For each variable $z$ in $(\{\mathbf{x}_i^{t}\},\{\mathbf{y}^{t}_i\}, \{\lambda_{i,l}^{t}\},\{\boldsymbol{\theta}_{i,j}^{t}\})$, we introduce a virtual learning rate as 
\begin{equation}
\setlength{\abovedisplayskip}{2pt}
\setlength{\belowdisplayskip}{2pt}
\tilde{\eta}^t_{i,z}=\left\{\begin{array}{l}
\eta^t_{i,z},  i \in Q^{t+1} \\
0, i \notin Q^{t+1} 
\end{array}\right..
\label{virtual}
\end{equation}

% \vspace{-2cm}
The expected learning rates are equalized to ensure the convergence of the algorithm. That is, if $\mathbb{E}[\eta^t_{i,z}] = \eta_z$, $\forall i$, the stationary points are maintained in expectation \cite{jeong2022asynchronous} as
\begin{equation}
\setlength{\abovedisplayskip}{2pt}
\setlength{\belowdisplayskip}{2pt}
\begin{aligned}
&\sum\limits_{i=1}^{N} \mathbb{E}[\tilde{\eta}^t_{i,z}]\nabla_{z}L'_{pi}(\{\mathbf{x}_i^{t}\},\{\mathbf{y}^{t}_i\}, \{\lambda_{i,l}^{t}\},\{\boldsymbol{\theta}_{i,j}^{t}\}) = 0\\
\Rightarrow & \sum\limits_{i=1}^{N} \nabla_{z} L'_{pi}(\{\mathbf{x}_i^{t}\},\{\mathbf{y}^{t}_i\}, \{\lambda_{i,l}^{t}\},\{\boldsymbol{\theta}_{i,j}^{t}\}) = 0.
\end{aligned}
\label{virtual2}
\end{equation}

% 为了统一各变量，都采取求和而不是平均的形式
\emph{Definition 3.} (Convergence metric) Inspired by \cite{mancino2022proximal, jiao2022asynchronous}, we use the a convergence metric with the stationary gap $\mathcal{G}^{t}$ and the consensus error $\mathcal{C}^t$ as follows:
%\begin{small}
%\vspace{-0.5em}
\begin{equation}
\setlength{\abovedisplayskip}{2pt}
\setlength{\belowdisplayskip}{2pt}
\Psi^t \triangleq ||\mathcal{G}^{t}||^2 + L^2\mathcal{C}^t. 
\label{eq31}
\end{equation}
%\end{small}
% The first term quantifies the convergence of variables to a stationary gap of the global objective, while the second term measures the consensus errors.

Specifically, the stationary gap is defined as
\begin{equation}
\setlength{\abovedisplayskip}{2pt}
\setlength{\belowdisplayskip}{2pt}
\mathcal{G}^{t}=\left[\begin{aligned}
&\{P(\mathbf{x}_{i}, \bar{\nabla}_{\mathbf{x}} L_{p}'(\{\mathbf{x}_i^{t}\},\{\mathbf{y}^{t}_i\}, \{\lambda_{i,l}^{t}\},\{\boldsymbol{\theta}_{i,j}^{t}\}), \tilde{\eta}_{i,x}^t)\}\\
&\{\bar{\nabla}_{\mathbf{y}} L_{p}(\{\mathbf{x}_i^{t}\},\{\mathbf{y}^{t}_i\}, \{\lambda_{i,l}^{t}\},\{\boldsymbol{\theta}_{i,j}^{t}\})\}\\
&\left\{\nabla_{\lambda_{i,l}} L_{p}(\{\mathbf{x}_i^{t}\},\{\mathbf{y}^{t}_i\}, \{\lambda_{i,l}^{t}\},\{\boldsymbol{\theta}_{i,j}^{t}\})\right\} \\
&\left\{\nabla_{\boldsymbol{\theta}_{i,j}} L_{p}(\{\mathbf{x}_i^{t}\},\{\mathbf{y}^{t}_i\}, \{\lambda_{i,l}^{t}\},\{\boldsymbol{\theta}_{i,j}^{t}\})\right\}
\end{aligned}\right],
\label{eq_32}
\end{equation}
where 
\begin{equation*}
\setlength{\abovedisplayskip}{2pt}
\setlength{\belowdisplayskip}{2pt}
\begin{array}{cc}
     &  \bar{\nabla}_{\mathbf{x}} L_{p}'(\cdot) \triangleq \frac{1}{N} \sum_{i=1}^{N}\nabla_{\mathbf{x}_{i}}L_{pi}'(\cdot), \\
     & \bar{\nabla}_{\mathbf{y}}L_{p}(\cdot) \triangleq   \frac{1}{N}\sum_{i=1}^{N}\nabla_{\mathbf{y}_i} L_{pi}(\cdot).
\end{array}
\end{equation*}

The consensus error is defined as  
\begin{equation}
\setlength{\abovedisplayskip}{2pt}
\setlength{\belowdisplayskip}{2pt}
\mathcal{C}^t = \sum_{i=1}^{N}(||\mathbf{x}_i^t - \bar{\mathbf{x}}^t ||^2 + ||\mathbf{y}_i^t - \bar{\mathbf{y}}^t||^2),
%\label{eq33}
\end{equation}
where $\bar{\mathbf{x}}^t = \frac{1}{N}\sum\limits^{N}_{i=1}\mathbf{x}_i^t$, $\bar{\mathbf{y}}^t = \frac{1}{N}\sum\limits^{N}_{i=1}\mathbf{y}_i^t$.

% 收敛性
% 收敛性矩阵
\begin{figure}[t]
\centering
\includegraphics[scale=0.12]{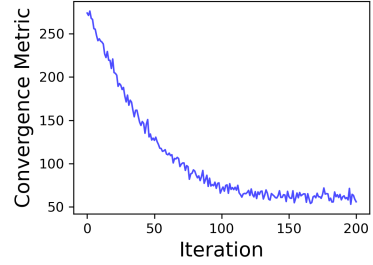}
%\vspace{-10mm}
\caption{Convergence performance of Argus.}
\label{fig:converge}
%\vspace{-5mm}
\end{figure}

\emph{Theorem 1.} ({Iteration Complexity}) Under Assumption 1 and 2, when step-sizes satisfy $\mathbb{E}[\eta_{i,x}^t] = \mathbb{E}[\eta_{i,y}^t] = \mathbb{E}[\eta_{i,\lambda}^t] = \mathbb{E}[\eta_{i,\theta}^t]$

\noindent $ = \eta$, and $\eta \le \{\frac{(1 - \rho)}{288 L N k_1 \tau}, \frac{\sqrt{1 - \rho}}{24 L \sqrt{N k_1 \tau}}, \frac{\sqrt{1 - \rho}}{24L \sqrt{2(N + M)(T_1 + T)}},\frac{1}{10 NL},$ 

\noindent $ \frac{(1 - \rho)^2}{640L^2N}, \frac{(1-\rho)^{\frac{3}{2}}}{40L}, \frac{1}{4L\sqrt{M}}\}$, then $(\{\mathbf{x}_i^{t}\},\{\mathbf{y}^{t}_i\}, \{\lambda_{i,l}^{t}\},\{\boldsymbol{\theta}_{i,j}^{t}\})$ generated by Argus satisfies:

% \newpage
\begin{small}
%\begin{small}
\begin{equation}
\setlength{\abovedisplayskip}{2pt}
\setlength{\belowdisplayskip}{2pt}
\begin{aligned}
& \frac{1}{T-1}\sum_{t=T_1+2}^{T_1+T} \mathbb{E}[\Psi^t] \\
\le & \frac{p}{T-1} (F^{T_1+2}- \underline{L}  +\frac{MNc_1^{1} \alpha_{1}}{2} + \frac{N^2c_2^{1} \alpha_{2}}{2} +\frac{4MN}{\eta} (\frac{c_{1}^{0}}{c_{1}^{1}} + \frac{c_{1}^{1}}{c_{1}^{2}})\alpha_{1} \\
+ & \frac{4N^2}{\eta} (\frac{c_{2}^{0}}{c_{2}^{1}} + \frac{c_{2}^{1}}{c_{2}^{2}}) \alpha_{2} +  4MN(c_1^{1})^2\alpha_1 + 4N^2(c_2^{1})^2\alpha_2 + \frac{6MN}{\eta}\sigma_{1}^{2} \\
+ & \frac{6N^2}{\eta} {\sigma_{2}}^{2} + \frac{c_{1}^{2}MN}{2}\alpha_{1}+\frac{c_{2}^{2}N^2}{2}\alpha_{2} ) = \mathcal{O}(\frac{1}{T-1}  ),
% \\= &  \mathcal{O}(\frac{1}{T-1}  ).
\end{aligned}
\label{eq35}
\end{equation}
\end{small}
%\end{small}

\noindent where $\sigma_{1}, \sigma_{2}, p$ are constants. The values of constants and more details are given in {Appendix.A.}

\emph{Discussion:} We observe the impact of network connectivity, represented by the parameter $\rho$, on the algorithm's convergence rate. Specifically, smaller values of $\rho$ allow for larger learning rates while maintaining convergence, and result in lower errors after a fixed number of iterations. Thus, a smaller $\rho$, corresponding to higher network connectivity, accelerates convergence, which aligns with the intuition that increased connectivity facilitates faster information propagation and, consequently, faster convergence.

\emph{Theorem 2.} (Communication Complexity) The communication complexity of Argus consists of two components. The first involves the exchange of optimization variables ($\mathbf{x}_{i}$, $\mathbf{y}_{i}$, ${\lambda}_{i,l}$ and $\boldsymbol{\theta}_{i,j}$) between each agent and its neighbors during each iteration, denoted by $\sum_{t=1}^T C_1^t$. The second component arises from the variables (${\mathbf{y}'_{i}}$, $\boldsymbol{\varphi}_{ij}$, $\boldsymbol{{\mathcal{P}}}_i$, and $\{\lambda_{i}\}$) exchanged during the periodic update of the cutting planes, which occurs every $\iota$ iterations up to the $T_1^{th}$ iteration, denoted by $C_2$. Specifically, Argus requires $\mathcal{O}(\sum_{t=1}^TC_1^t + C_2)$ in communication complexity, where $C_1^t = 32d^t(N(m+n) + \sum_{i=1}^Np_i(|\mathcal{P}_i^t|+nd^t))$, $C_2 = 32\sum_{t\in \mathcal{T}}(Nd^t(K(m+md^t)+(d^t(n+m) + 1)))$, $d^t = \frac{1}{N}\sum_{i=1}^{N}\sum_{j=1}^{N}\mathbf{W}^t$, and $\mathcal{T} = \{\iota, \cdots, \left\lfloor \frac{T_1}{\iota}\right\rfloor\} \cdot \iota$. More details are given in {Appendix.B}. 

% 计算复杂度
\emph{Theorem 3.} (Computational Complexity) The computational complexity of Argus consists of two components: the computation per iteration (Eq.(15) to Eq.(20)), denoted by $\sum_{t=1}^T C_{P_1}^t$, and the computation involved in updating the cutting planes (Eq.(5), Eq.(6), Eq.(24) to Eq.(26)), denoted by $C_{P_2}$. Specifically, Argus requires $\sum_{t=1}^TC_{P_{1}}^t + C_{P_{2}}$ in communication complexity, where $C_{P_{1}}^t = \mathcal{O}(N\left|\mathcal{P}_{i}^{t}\right|^{2} d^{t}(n+m))+\mathcal{O}(N d^{t^{2}} n)$, $C_{P_{2}} = \sum_{t\in \mathcal{T}}(\mathcal{O}(Nd^t(n+m) + NmK))$. More details are given in {Appendix.C}.

\section{Experiments} 
\label{experiments}
In this section, we conduct experiments to address the following questions:
% 1.在同步设置下，所提出的方法是否比其他去中心化双层优化基线算法性能更优越？
% 2.所提出方法的异步版本是否比同步版本算法收敛更快？
% 3.所提出方法的核心部件是否在取得好的性能上是不可或缺的？（我们采用同步场景来消除通信的影响）
% 4.与不能利用卫星的算法相比，所提出的算法是否能够加快算法的收敛？利用卫星传输神经网络具有具有可行性吗？
\begin{itemize}
    \item Does the proposed method outperform other (synchronous) decentralized bilevel optimization algorithms in a synchronous setting?
    \item Does Argus converge faster than its synchronous version (Argus-S)?
    \item Are the core components of the proposed method indispensable in achieving good performance? 
\end{itemize}

\subsection{Experimental Setup}

\emph{Tasks and Datasets.} We examine Argus in the context of two machine learning tasks: meta-learning and hyperparameter optimization, and also explore a specialized application scenario in disaster response.

% 元学习
\begin{figure*}%[htbp]
\vspace{-0.02\textwidth}
\centering
\subfigure[Omniglot]{
\includegraphics[width=0.45\linewidth]{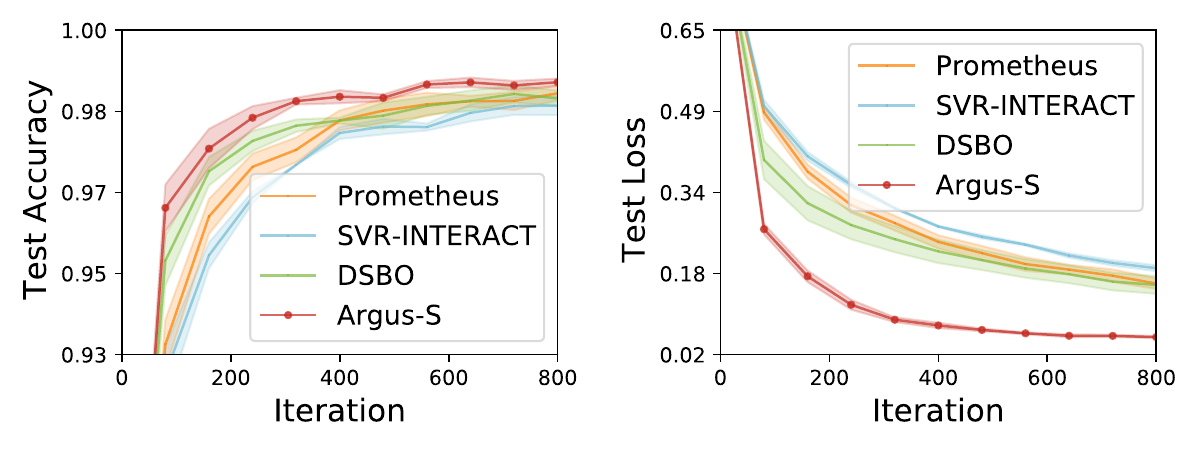}}
\hspace{20pt}
% fig 2
\subfigure[Double MNIST]{
\includegraphics[width=0.45\linewidth]{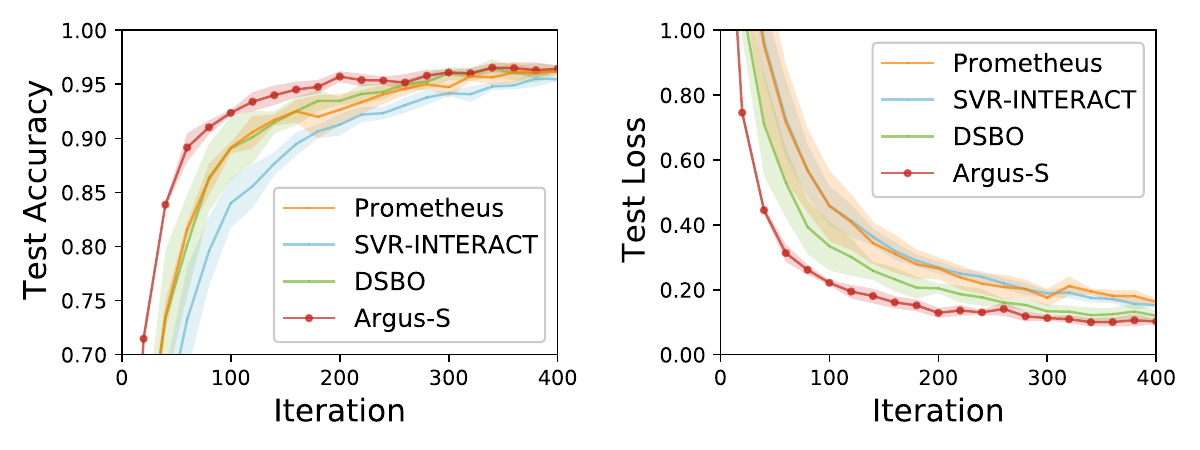}}
\vspace{-2pt}
\vspace{-0.01\textwidth}
\caption{Performance comparison with baseline methods on meta-learning.} %  %大图名称
\label{fig:result_meta}  %图片引用标记
\end{figure*}

% 超参数优化
\begin{figure*}%[htbp]
\vspace{-0.05\textwidth}
\centering
\subfigure[MNIST]{
\includegraphics[width=0.45\linewidth]{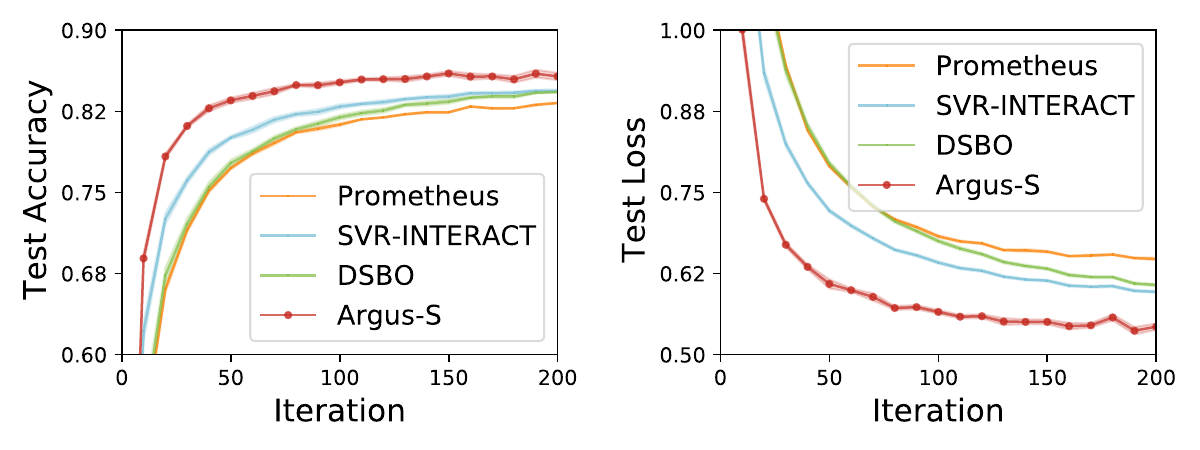}}
\hspace{20pt}
\subfigure[Fashion MNIST]{
\includegraphics[width=0.45\linewidth]{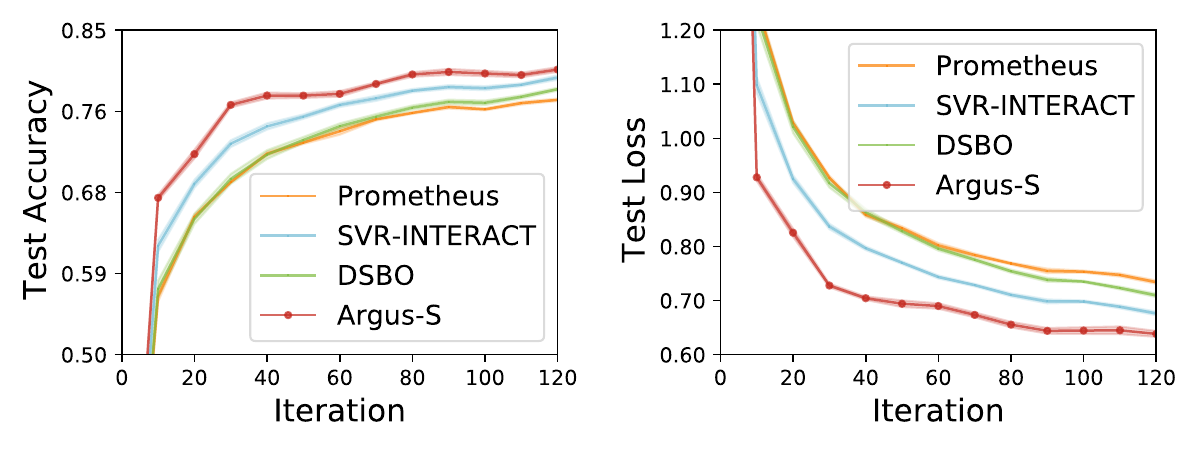}}
% fig 2
\vspace{-2pt}
\vspace{-0.01\textwidth}
\caption{Performance comparison with baseline methods on hyperparameter optimization.} %  %大图名称
\label{fig:result_hyper}  %图片引用标记
\end{figure*}

%灾难响应
\begin{figure*}%[htbp]
\vspace{-0.05\textwidth}
\centering
\subfigure[AIDER]{
\includegraphics[width=0.45\linewidth]{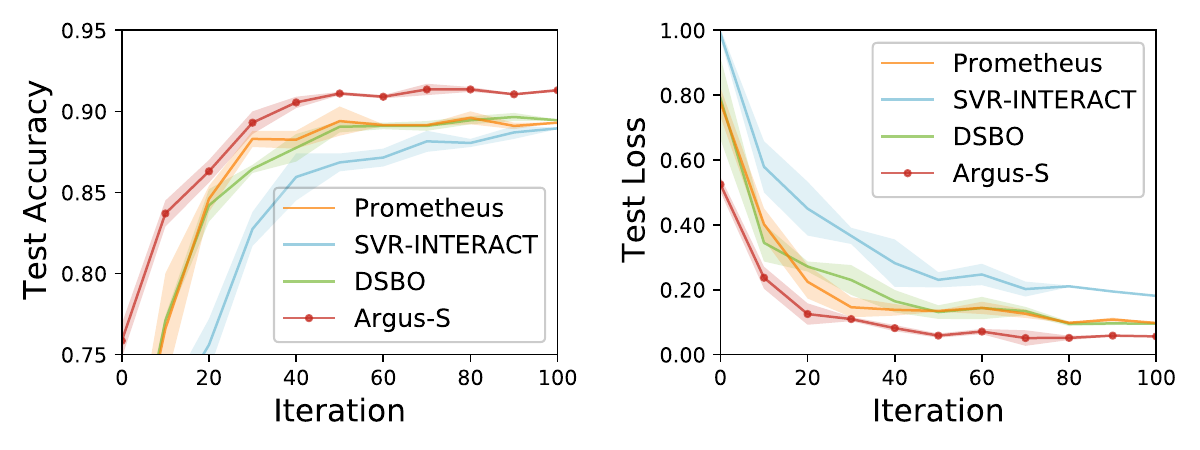}}
\hspace{20pt}
\subfigure[MEDIC]{
\includegraphics[width=0.45\linewidth]{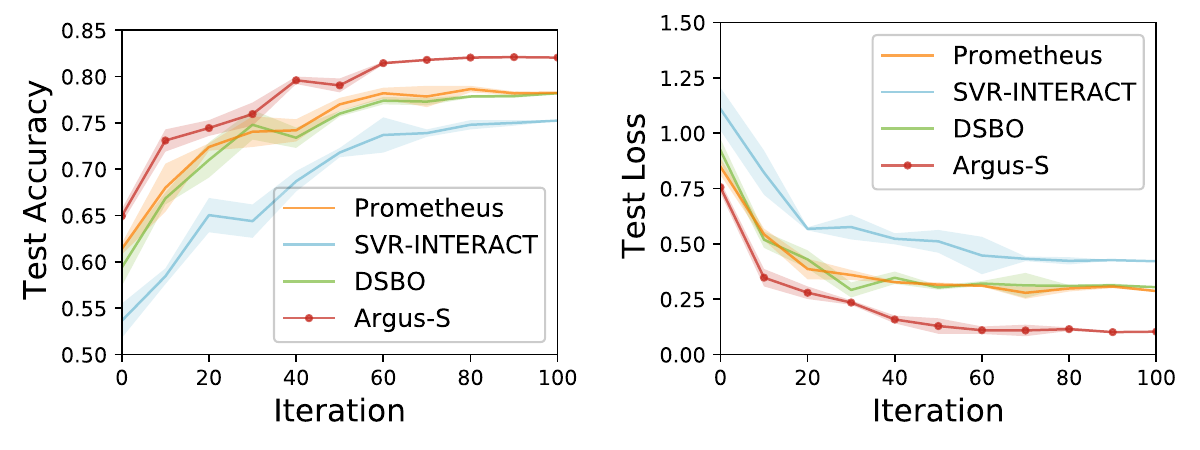}}
% fig 2
\vspace{-2pt}
\vspace{-0.01\textwidth}
\caption{Performance comparison with baseline methods on continual learning.} %  %大图名称
\label{fig:result_continual}  %图片引用标记
\end{figure*}

% ---------------- 消融实验 ----------------
% 元学习
\begin{figure*}%[htbp]
\vspace{-0.05\textwidth}
\centering
\subfigure[Omniglot]{
\includegraphics[width=0.45\linewidth]{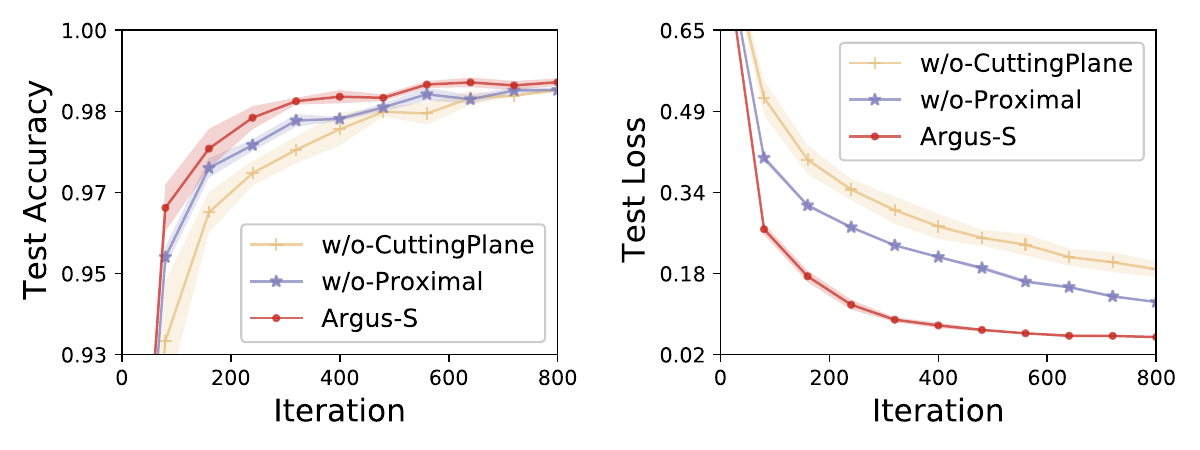}}
\hspace{20pt}
% fig 2
\subfigure[Double MNIST]{
\includegraphics[width=0.45\linewidth]{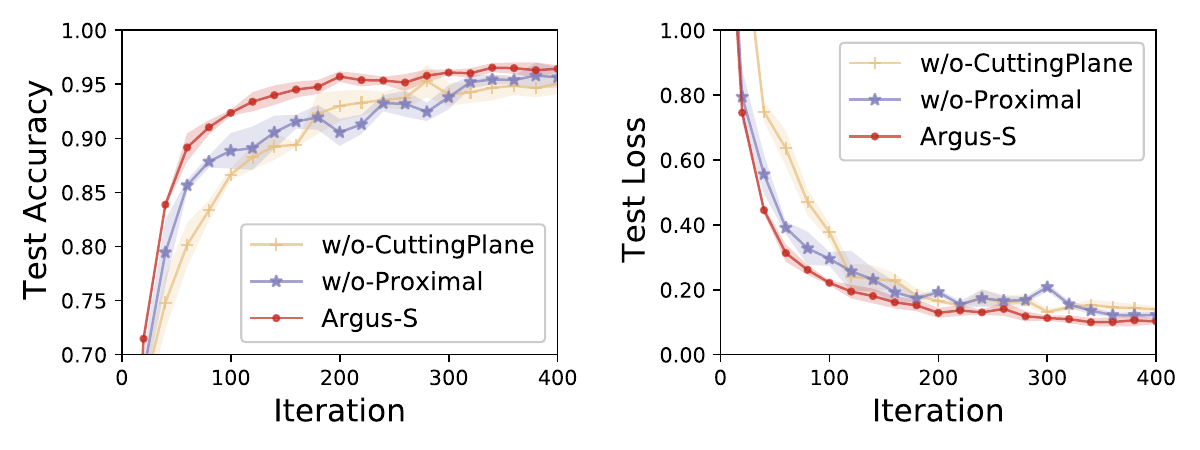}}
\vspace{-2pt}
\vspace{-0.01\textwidth}
\caption{Results of ablation experiments on meta-learning.} %  %大图名称
\label{fig:abla_meta}  %图片引用标记
\end{figure*}

% 超参数优化
\begin{figure*}%[htbp]
\vspace{-0.05\textwidth}
\centering
\subfigure[MNIST]{
\includegraphics[width=0.45\linewidth]{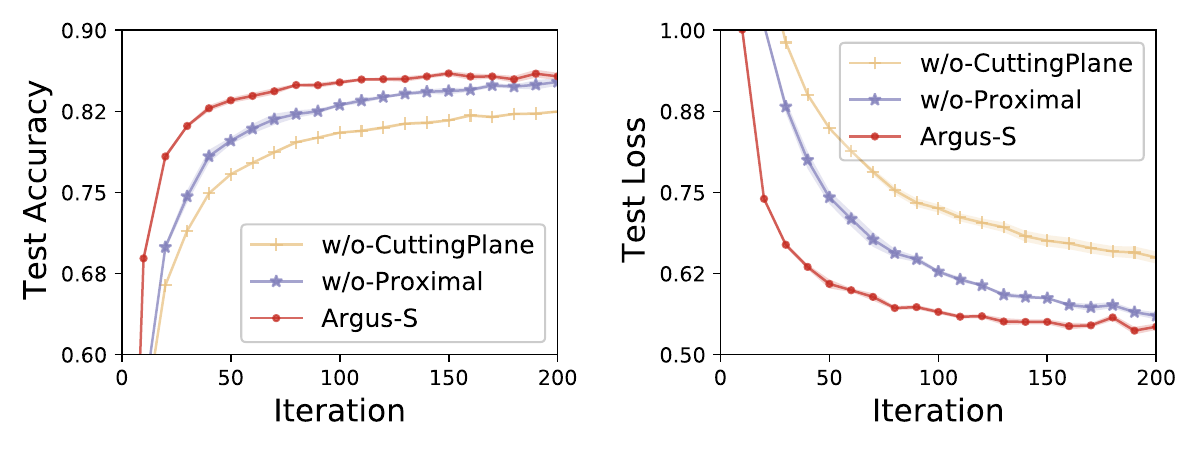}}
\hspace{20pt}
\subfigure[Fashion MNIST]{
\includegraphics[width=0.45\linewidth]{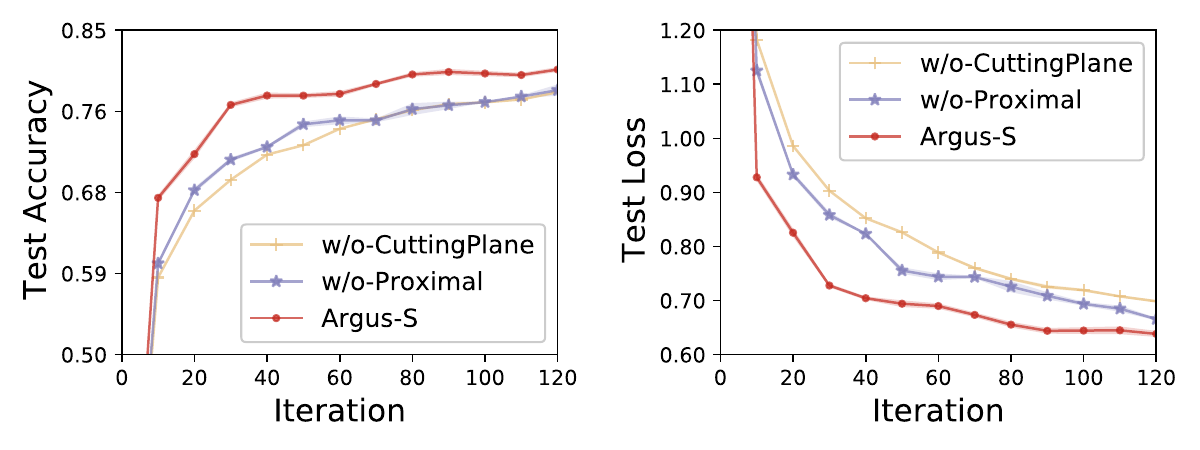}}
% fig 2
\vspace{-2pt}
\vspace{-0.01\textwidth}
\caption{Results of ablation experiments on hyperparameter optimization.} %  %大图名称
\label{fig:abla_hyper}  %图片引用标记
\end{figure*}

% 灾难响应
\begin{figure*}[htbp]
\vspace{-0.05\textwidth}
\centering
\subfigure[AIDER]{
\includegraphics[width=0.45\linewidth]{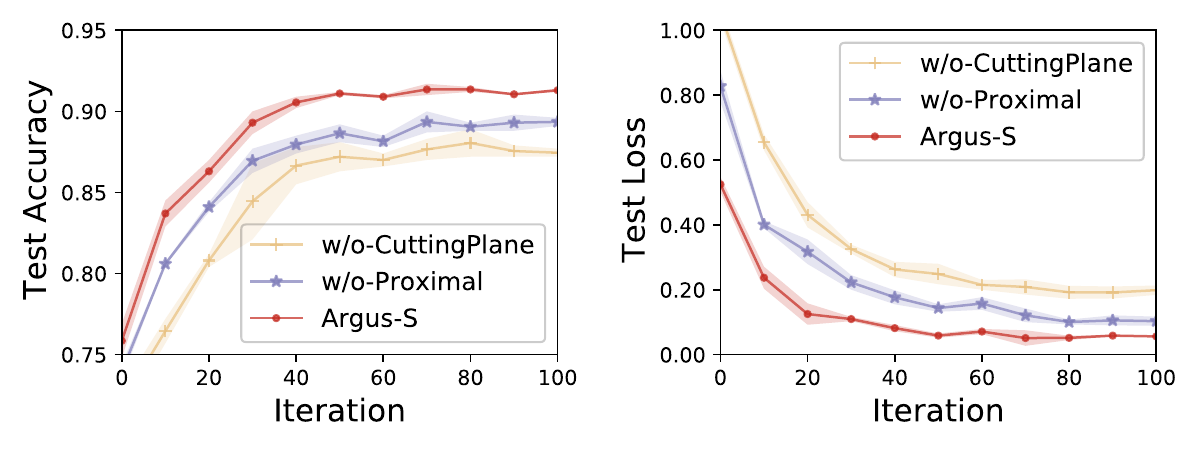}}
\hspace{20pt}
\subfigure[MEDIC]{
\includegraphics[width=0.45\linewidth]{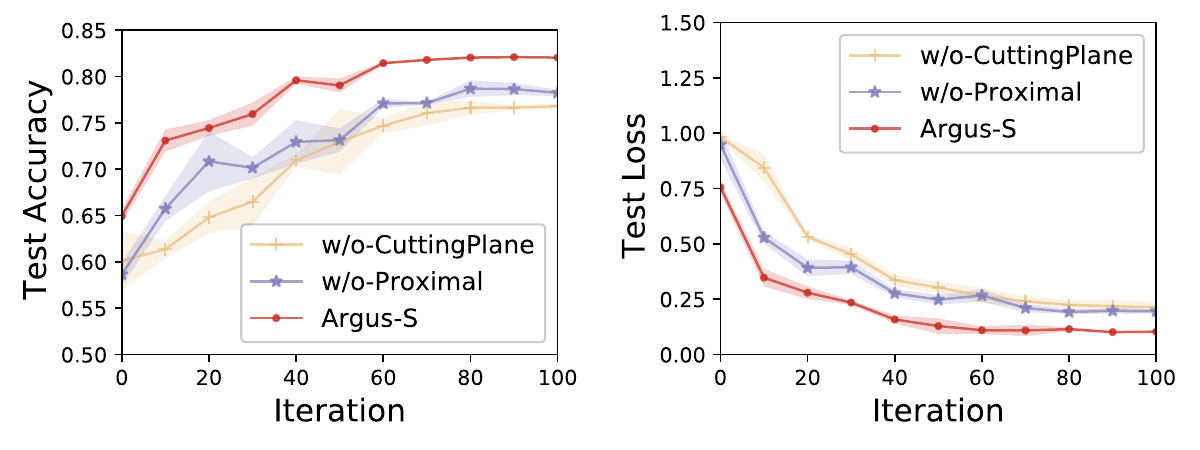}}
% fig 2
\vspace{-2pt}
\vspace{-0.01\textwidth}
\caption{Results of ablation experiments on continual learning.} %  %大图名称
\label{fig:abla_continual}  %图片引用标记
\end{figure*}

(1) Meta-Learning. We consider the meta-learning problem in the context of few-shot supervised learning. The decentralized meta-learning problem in our experiments is:
%\begin{small}
\begin{equation}
%\vspace{-0.5em}
\setlength{\abovedisplayskip}{2pt}
\setlength{\belowdisplayskip}{2pt}
\begin{aligned}
\underset{
\boldsymbol{\theta}}{\min} & \frac{1}{N}\underset{i=1}{\overset{N}{\sum}} F_i(\boldsymbol{\theta}_i) \triangleq \mathcal{L}(\boldsymbol{\varphi}_i(\boldsymbol{\theta}_i);\mathcal{D}_i^{val}) + R(\boldsymbol{\theta}_i)\\ 
\text{s.t.} & \text{ } \boldsymbol{\varphi}(\boldsymbol{\theta}) = \underset{\boldsymbol{\varphi}'}{\arg\min} \text{ }  \frac{1}{N}\underset{i=1}{\overset{N}{\sum}} f_i(\boldsymbol{\theta}_i, \boldsymbol{\varphi}_i') \\ &  f_i(\boldsymbol{\theta}_i, \boldsymbol{\varphi}_i') \triangleq \mathcal{L}(\boldsymbol{\varphi}_i';\mathcal{D}_i^{tr}) +
r(\boldsymbol{\varphi}_i')+||\boldsymbol{\varphi}_i'-\boldsymbol{\theta}_i||^2,
\end{aligned}
\label{eq36}
\end{equation}
%\end{small}
where $\boldsymbol{\theta}_i$ and $\boldsymbol{\varphi}_i$ denote the meta-parameter and task-specific parameter of the $i^{th}$ agent, respectively. $\mathcal{D}_i^{tr}$ and $\mathcal{D}_i^{val}$ are the training and validation datasets on the $i^{th}$ agent. $N$ is the number of agents, $\mathcal{L}$ is the cross-entropy loss, $r$ and $R$ are $l_1$ norm.

(2) Hyperparameter Optimization. We consider the hyperparameter optimization in the context of data hyper-cleaning \cite{franceschi2017forward}. It involves training a model in contaminated environments where each training label is changed randomly. We discuss the following decentralized data hyper-cleaning problem:
%\newpage
%\begin{small}
\begin{equation}
\setlength{\abovedisplayskip}{2pt}
\setlength{\belowdisplayskip}{2pt}
\begin{aligned}
\underset{
\boldsymbol{\psi }}{\min} & \frac{1}{N}\underset{i=1}{\overset{N}{\sum}} F_i(\boldsymbol{\psi }_i,\boldsymbol{w}_i) \triangleq \frac{1}{\vert \mathcal{D}_{i}^{val}\vert} \underset{(\mathbf{x}_{j},y_{j})\in \mathcal{D}_{i}^{val}}{\sum}\mathcal{L}(\mathbf{x}_j^{\top}\boldsymbol{w}_i,{y}_j)\\ 
\text{s.t.} & \text{ } \boldsymbol{w} = \underset{\boldsymbol{w}'}{\arg\min} \text{ }  \frac{1}{N}\underset{i=1}{\overset{N}{\sum}} f_i(\boldsymbol{\psi }_i,\boldsymbol{w}_i')\\
& f_i(\boldsymbol{\psi }_i,\boldsymbol{w}_i') \triangleq \frac{1}{\vert \mathcal{D}_{i}^{tr}\vert} \underset{(\mathbf{x}_{j},y_{j})\in \mathcal{D}_{i}^{tr}}{\sum}\sigma ({\psi }_{i,j})\mathcal{L}(\mathbf{x}_j^{\top}\boldsymbol{w}_i',{y}_j)+R(\boldsymbol{w}_i'),
\end{aligned}
\label{eq37}
\end{equation}
%\end{small}
\noindent where $\mathcal{D}_i^{tr}$ and $\mathcal{D}_i^{val}$ are the training and validation datasets on the $i^{th}$ agent. $(\mathbf{x}_{j},y_{j})$ denotes the $j^{th}$ data and label. $\sigma$ represents the sigmoid function, $\mathcal{L}$ is the cross-entropy loss, $R$ is $l_1$ norm.

% 新实验
(3) Disaster Response.
We consider a problem of continual learning for disaster image classification. The specific formulation of this problem is as follows:
\begin{equation}
\setlength{\abovedisplayskip}{2pt}
\setlength{\belowdisplayskip}{2pt}
\begin{aligned}
\underset{\mathbf{x}}{\min} & \  \underset{i=1}{\overset{N}{\sum}} F_i(\mathbf{x}_i, \mathbf{y}_i) \triangleq \mathcal{L}_u(\mathbf{x}_i, \mathbf{y}_i) + r(\mathbf{x}_i)\\ 
\text{s.t.} &\ \mathbf{y} = \underset{\mathbf{y}'}{\arg\min} \text{ } \underset{i=1}{\overset{N}{\sum}} f_i(\mathbf{x}_i, \mathbf{y}_i') \triangleq \mathcal{L}_l(\mathbf{y}_i') + R(\mathbf{x}_i,\mathbf{y}_i') + r(\mathbf{y}_i'),
\end{aligned}
\label{re_1}
\end{equation}
where $\mathbf{x}$ denotes the historical model parameters and $\mathbf{y}$ denotes the current model parameters. The lower-level problem optimizes the current parameters for the new task using the classification loss function $\mathcal{L}_l(\cdot)$, with the $l_2$ norm regularization term $R(\cdot)$ constraining the similarity between the updated current and historical parameters. The upper-level problem ensures performance stability between the updated and historical models, where $\mathcal{L}_u(\cdot)$ is the cross-entropy between their outputs, and $r(\cdot)$ is the $l_1$ norm of the model parameters.

For meta-learning, we use the Omniglot \cite{vinyals2016matching} and Double MNIST \cite{morerio2017curriculum} datasets. For the hyper-parameter optimization task, we utilize MNIST \cite{lecun1998gradient} and Fashion MNIST \cite{xiao2017fashion} datasets. For disaster response, we conducted experiments on the AIDER \cite{9050881} and MEDIC \cite{alam2023medic} datasets. Our experiments involve a network of $10$ agents, generated using the Erdős-Rényi random graph approach with a connectivity probability $p_c = 0.5$. Each agent represents an AAV, and the network topology is envolving in each iteration to imitate the dynamics of SAGIN.

% 哪个实验结果可以说明可行性？It is proved that it is feasible and effective to use the proposed anomaly detection algorithm for training in SAGIN

% 3 同步异步对比实验
\begin{figure}[t]
% fig 1
\includegraphics[scale=0.4]{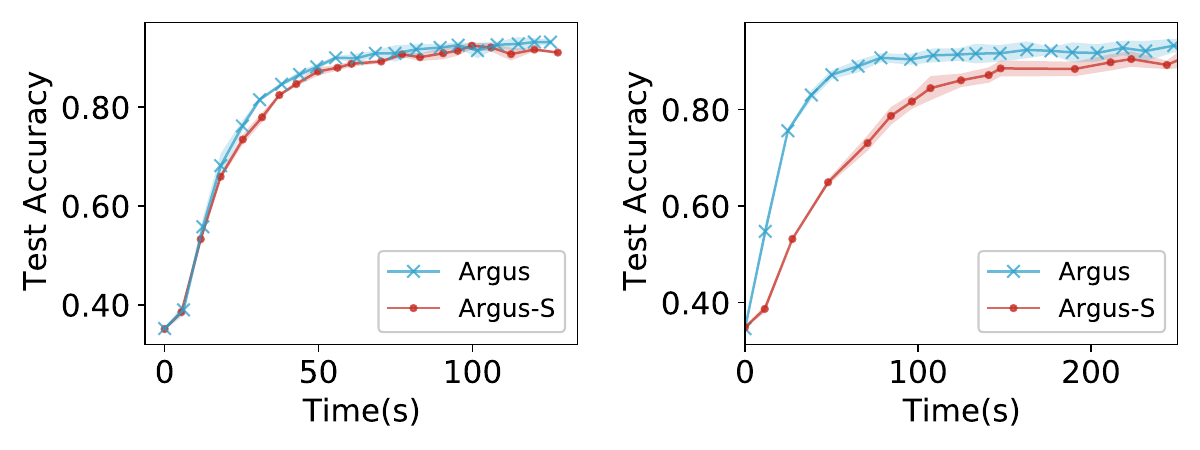}
\vspace{-2mm}
\caption{Performance comparison of synchronous (Argus-S) and asynchronous (Argus) methods. Left: Ideal Network Conditions. Right: Straggler-Inclusive Conditions.}
\label{fig:result3}  %图片引用标记
\end{figure}

% % 6
% \begin{figure}[t]
% \centering
% \includegraphics[scale=0.4]{fig/STRUG.pdf}
% %\vspace{-10mm}
% \caption{Comparison of the effects of using satellite communications.}
% \label{fig:result4}
% %\vspace{-5mm}
% \end{figure}

\emph{Baselines.} There is still a lack of asynchronous baselines for the non-convex and non-smooth decentralized federated bilevel optimization problem in time-varying networks. Therefore, we compare Argus with some of the most relevant algorithms for decentralized bilevel optimization: Prometheus \cite{liu2023prometheus}, SVR-INTERACT \cite{liu2022interact} and DSBO \cite{yang2022decentralized}. Since SVR-INTERACT and DSBO aim to solve smooth problems, we replace the gradient with a subgradient \cite{paszke2019pytorch} in these methods to deal with non-smooth terms in our experiments. All three baseline methods are synchronous algorithms proposed for non-convex upper-level problems and strongly-convex lower-level problems.

\subsection{Experiment Evaluation}

\textbf{Convergence Performance.} Using the convergence metric $\Psi$ as defined in Definition 3, we illustrate the convergence performance of Argus in Figure \ref{fig:converge} on the hyperparameter optimization task and the MNIST dataset. It is evident that Argus gradually converges with each iteration.

\textbf{Performance Comparison with Baselines.} Since all baseline methods are synchronous, we compare Argus-S, the synchronous version of Argus, with baseline methods under synchronous conditions to fairly demonstrate the superiority of the mechanisms in our algorithm other than asynchronous communication. In Argus-S, all agents participate in the computation during each iteration, that it, $\forall t, \mathcal{Q}^t = [N]$. The results of the meta-learning, hyperparameter optimization, and disaster response are shown in Figure \ref{fig:result_meta}, \ref{fig:result_hyper}, and \ref{fig:result_continual}, respectively. From the above results, we can draw the following conclusions: (1) The loss value of our algorithm (Argus-S) decreases faster with each iteration. (2) Our algorithm achieves higher final accuracy. (3) Our algorithm exhibits stronger performance stability. For instance, the variance in performance of our algorithm on Double MNIST dataset is significantly lower than that of the baseline methods.
% 随迭代轮次“收敛”的速度更快
% 最终的准确性更高
% 其他数据集上的性能方差类似。但在DM上性能方差比其他方法小。

\textbf{Ablation Experiments.} First, we present two stripped-down versions of Argus-S to verify the effectiveness of our core modules, namely proximal gradient and cutting planes. Specifically, \emph{w/o-CuttingPlane} iteratively updates the upper-level and lower-level variables of the original decentralized bilevel optimization problems in Eq.(\ref{eq1}) based on the proximal gradient descent. \emph{w/o-Proximal} replaces the proximal operator in Argus-S with the subgradient. Figure \ref{fig:abla_meta}, \ref{fig:abla_hyper}, and \ref{fig:abla_continual} demonstrate the efficiency and indispensability of the polyhedral approximation and the proximal operator. Moreover, since the \emph{w/o-Proximal} outperforms the \emph{w/o-CuttingPlane}, it can be seen that the cutting plane approximation is more impact on the performance of the algorithm than the proximal gradient. There are two main reasons why Argus-S performs better than any of its stripped-down versions: (1) The proximal component is effective in solving non-smooth objectives. (2) The proposed single-loop algorithm based on polyhedral approximation is both accurate and computationally efficient for non-convex functions.

% 异步和同步对比 
\textbf{Advantage of the Asynchronous Mechanism}. We compare Argus with its synchronous version (Argus-S) under two network conditions. The experiment assumes that the average time for an agent to complete an iteration is influenced by both computation and communication delays, which depend on the agent’s computational capabilities and network conditions. The first scenario, ``Ideal Network Conditions'', assumes minimal agent heterogeneity and optimal network performance, where agents are well-balanced and less likely to fall behind. In this setup, the average iteration time for each agent is drawn from the same distribution. The second scenario, ``Straggler-Inclusive Conditions'', simulates a situation with significant differences in agent capabilities or poor network conditions. Here, two agents are randomly selected as stragglers in each iteration, facing much higher computation or communication delays, and are more likely to miss global iterations.

Figure \ref{fig:result3} shows the results on the Double MNIST dataset. In the ``Ideal Network Conditions'' setup (left plot), Argus closely matches Argus-S, as most agents participate in the computation during nearly all rounds. As a result, there is no significant difference in training speed. In the ``Straggler-Inclusive Conditions'' (right plot), Argus clearly outperforms Argus-S in training speed, achieving higher accuracy within the same training time. This advantage stems from Argus’ ability to complete global iterations without waiting for stragglers, resulting in shorter iteration times. These results highlight Argus’ substantial advantage in scenarios with stragglers, which better mirror real-world network conditions.

\begin{figure}[htbp]
\vspace{-0.02\textwidth}
\hspace{-0.08\textwidth}
    \centering
    \begin{minipage}[b]{0.18\textwidth} 
        \centering
        \includegraphics[width=1.3\textwidth]{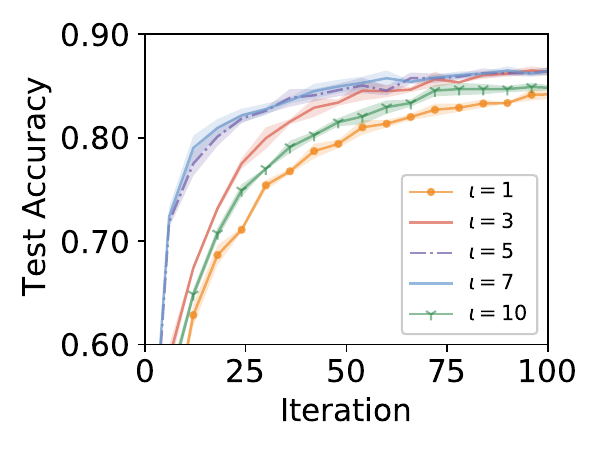}
        \vspace{-0.021\textwidth}
        \caption{{Impact of $\iota$ on test accuracy.}}
        \label{fig_iota}
    \end{minipage}
    \hspace{0.05\textwidth} % 可调整间距
    \begin{minipage}[b]{0.18\textwidth} 
        \centering
        \includegraphics[width=1.3\textwidth]{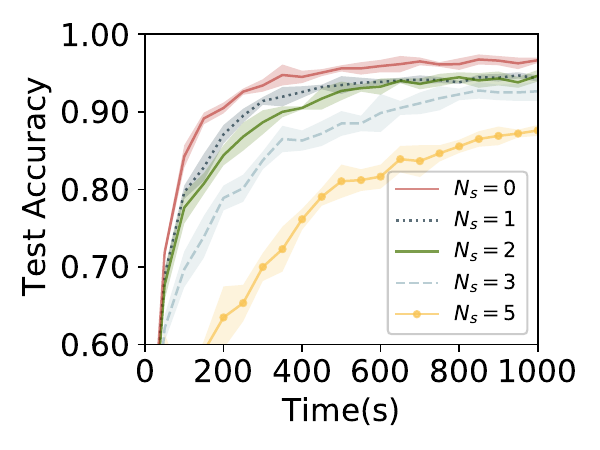}
        \caption{{Test accuracy with varying low-performance agents.}}
        \label{fig_heto}
    \end{minipage}
\end{figure}

{
\textbf{Impact of the Update Frequency of the Cutting Planes}. We conducted an experiment on the MNIST dataset to investigate the effect of the update frequency of the cutting planes, denoted as $\iota$, on algorithm performance. As shown in Figure \ref{fig_iota}, the convergence rate does not vary significantly when $\iota$ is set to 5 or 7. However, when $\iota$ is smaller (e.g., 3 or 1), the convergence rate decreases. For larger values of $\iota$ (e.g., 10), the convergence rate is also slower than when $\iota$ is 5 or 7. In summary, a value of $\iota$ that is too large slows convergence due to lower solution quality, while a value that is too small reduces convergence speed because of the high computational cost of the cutting planes. In practice, the suitable value of $\iota$ may vary across different datasets. We do not recommend using excessively small values of $\iota$, as they directly impact computational complexity.}

{
% 异构场景
\textbf{Performance in Heterogeneous Networks}. We designed an experiment to evaluate the performance of Argus in heterogeneous networks. The experimental setup consisted of a heterogeneous device cluster, with Docker deployed on four PCs and 10 virtual containers serving as agents. These agents were categorized into two performance tiers: high-performance (1 CPU core, 4GB memory) and low-performance (0.5 CPU core, 2GB memory). The low-performance agents experienced frequent memory swapping and slower computation speeds, resulting in increased iteration times and a higher likelihood of becoming stragglers. The algorithm's performance was evaluated under different numbers of low-performance agents ($N_s$ = 0, 1, 2, 3, 5).}

{
The results on the Double MNIST dataset are presented in Figure \ref{fig_heto}. As shown, the convergence rate generally decreases as the number of low-performance agents increases. When $N_s=1$ or $N_s=2$, convergence remains relatively fast and accuracy remains high, indicating that the algorithm is robust to the presence of straggler nodes in heterogeneous networks. However, when $N_s=5$, convergence slows significantly, and accuracy stabilizes at a lower value, highlighting that an excessive number of stragglers disrupts the aggregation process. This disruption hampers both convergence speed and model accuracy, as fewer agents contribute to the process and increased staleness within the network undermines the consistency of information aggregation. These results suggest that while Argus is somewhat robust to stragglers, a high proportion of stragglers severely degrades performance. The algorithm must strike a balance between training efficiency and accuracy: waiting for more agents to update may improve performance, but in highly heterogeneous environments, this can lead to inefficiency and unpredictable delays.}

\section{Conclusion}
% ** 需要修改 卫星的描述
\label{conclusion}
We propose Argus, a novel asynchronous algorithm for decentralized federated non-convex bilevel learning over 6G SAGIN. Argus enables asynchronous agent aggregation in time-varying networks, mitigating the impact of stragglers on the training speed. Theoretical analyses affirm that Argus achieves a {iteration complexity} of $\mathcal{O}(1/\epsilon)$. Our empirical studies in the context of meta-learning, hyperparameter optimization and continual learning validate the effectiveness of Argus. To the best of our knowledge, this work represents a pioneering endeavor to explore bilevel learning over 6G SAGIN.

\section{Future Directions}
\label{future}
In advancing decentralized federated learning (DFL) within 6G Space-Air-Ground Integrated Networks (SAGIN), critical challenges persist, especially in  privacy protection, communication efficiency, and multimodal data integration. {In Argus, the exchange of local parameters between agents may pose privacy risks, such as model inversion attacks. To enhance privacy, techniques such as homomorphic encryption (HE) and differential privacy (DP) can be used. The Argus algorithm's convergence is guaranteed by the cutting plane method, where infeasible solutions are excluded as the number of cutting planes increases, and the approximated solution space approaches the feasible domain. Therefore, introducing secure aggregation does not affect convergence. However, both HE and DP in DFL can slow the convergence rate due to their computational overhead, with HE being particularly costly due to the complex protocols required for agent interactions. This creates a trade-off between privacy and speed. In the future, Argus can securely aggregate model parameters while using standard mechanisms for dual variables and cutting plane parameters to minimize computational costs, thus balancing privacy and performance.} Besides, high inter-device message exchanges in DFL can lead to communication bottlenecks in large-scale networks like SAGIN, where resources are limited. As network scale grows, the communication complexity of our algorithm increases, a common issue in decentralized systems. One potential solution is to reduce the number of communicating neighbors for each agent. Over-the-air (OTA) computation can alleviate resource constraints by enabling shared radio resources for model updates. Additionally, this study assumes homogeneous data across agents, but SAGIN's heterogeneous environment requires methods to handle multimodal data from satellite, aerial, and ground sensors. Heterogeneous graph neural networks (GNNs) are well-suited for this, given the diverse node types and complex relationships. However, multimodal learning often requires large labeled datasets, which are scarce in SAGIN due to high labeling costs, such as for remote sensing and specialized intelligence. Thus, exploring unsupervised and transfer learning approaches is crucial for improving model generalization in these contexts.

\bibliographystyle{IEEEtran}
\normalem
\bibliography{myreference}

% \vspace{-1cm}

\begin{IEEEbiography}[{\includegraphics[width=1in,height=1.25in,clip,keepaspectratio]{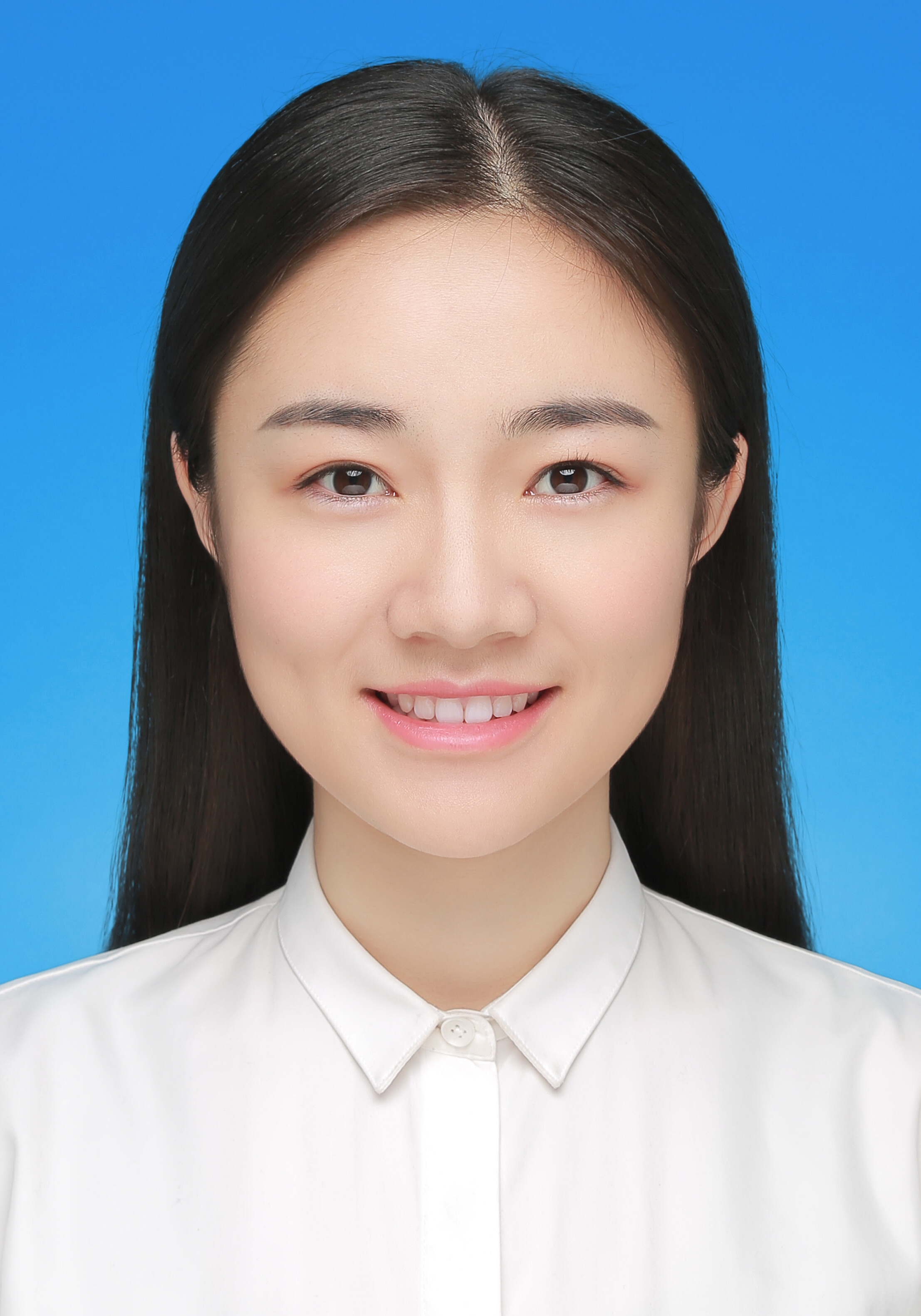}}]{Ya Liu} received the B.S degree from Beijing Jiaotong University, Beijing, China, in 2019. She is currently pursuing the Ph.D. degree in computer science from the Department of Computer Science, Tongji University, Shanghai, China. Her current research interests include anomaly detection, network security, and distributed learning.
\end{IEEEbiography}

% \newpage
\vspace{-1cm}

\begin{IEEEbiography}[{\includegraphics[width=1in,height=1.25in,clip,keepaspectratio]{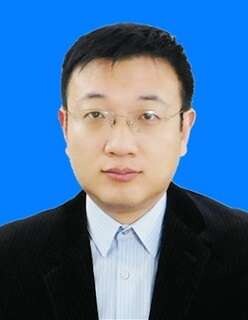}}]{Kai Yang} (Senior Member, IEEE) received the Ph.D. degree from Columbia University, USA. He has held various prestigious positions, including Technical Staff Member at Bell Laboratories, Alcatel-Lucent, USA. From 2011 to 2016, he has also served as an adjunct faculty member at Columbia University. 

Currently, he is a Distinguished Professor at Tongji University, Shanghai, China. He holds over thirty patents and has been extensively published in Top AI conferences and leading IEEE journals and conferences. His current research interests encompass AI+Networking, AI for Time Series, Nested Optimization, and Wireless Sensing and Communications. He has also served as a Technical Program Committee (TPC) member for numerous AI and IEEE conferences. He has received several prestigious awards, including the Eliahu Jury Award from Columbia University, the Bell Laboratories Teamwork Award, and the Best Paper Honorable Mention Award for time series workshop of IJCAI 2023. From 2012 to 2014, he was the Vice-Chair of the IEEE ComSoc Multimedia Communications Technical Committee. Additionally, he has held leadership roles such as Demo/Poster Co-Chair for IEEE INFOCOM, Symposium Co-Chair for IEEE GLOBECOM, and Workshop Co-Chair for IEEE ICME.
\end{IEEEbiography}

\vspace{-1cm}

\begin{IEEEbiography}[{\includegraphics[width=1in,height=1.25in,clip,keepaspectratio]{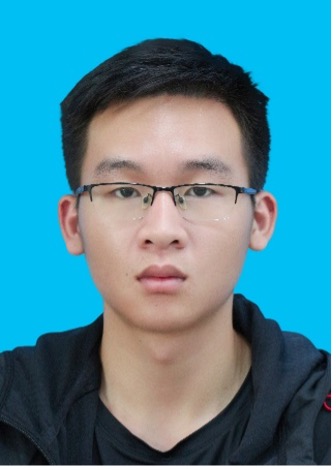}}]{Yu Zhu} received the B.E degree from Jiangnan University, Wuxi, China in 2023. He is currently pursuing the M.E degree in computer science from the Department of Computer Science, Tongji University, Shanghai, China. His current research interests include distributed learning and prompt engineering.
\end{IEEEbiography}

\vspace{-1cm}

\begin{IEEEbiography}[{\includegraphics[width=1in,height=1.25in,clip,keepaspectratio]{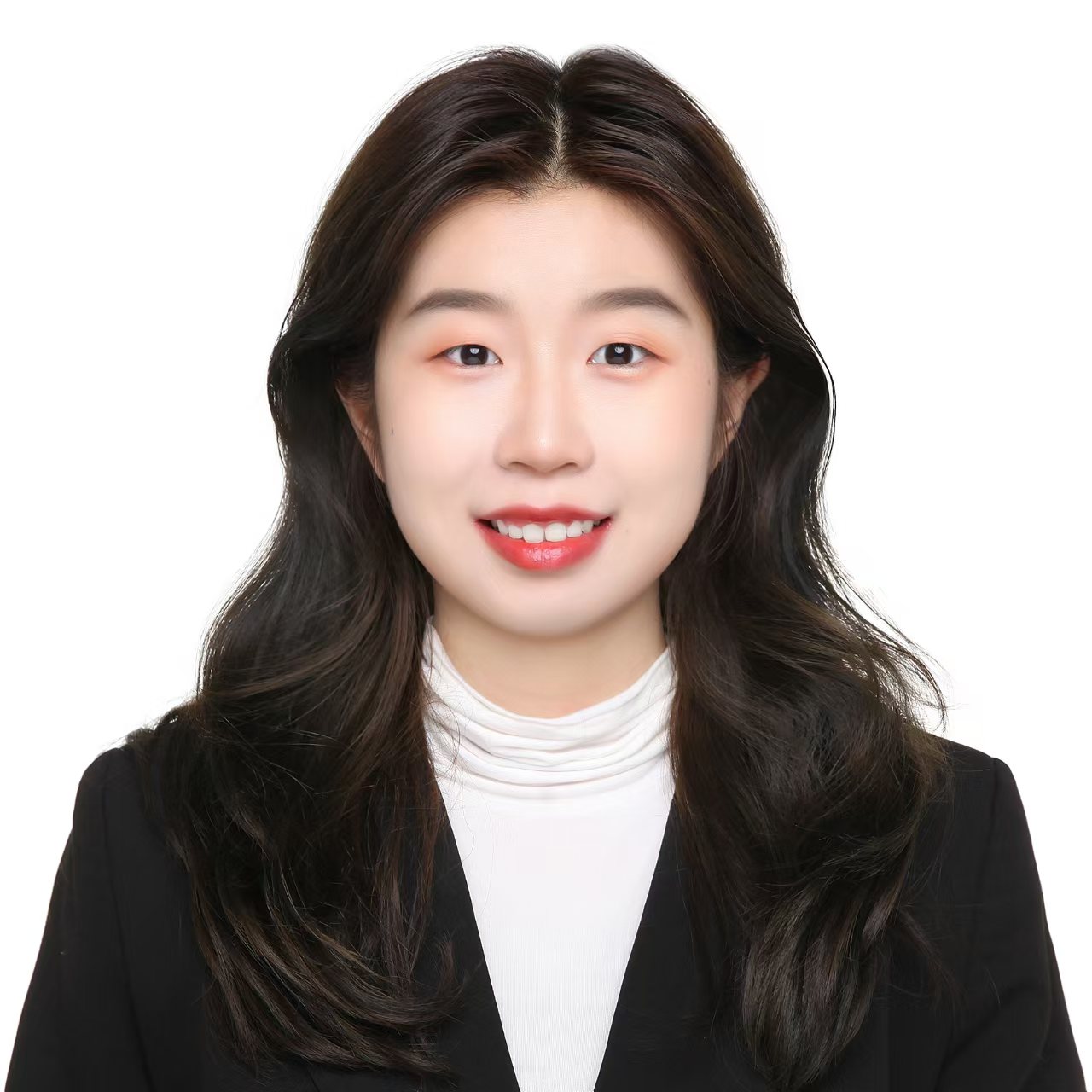}}]{Keying Yang} received the B.S degree from Tongji University, Shanghai,China,in 2023. Currectly,she is pursuing the Ph.D. degree in computer science from the Department of Computer Science, Tongji University, Shanghai, China. Her current research interests include LLM reasoning, prompt engineering.
\end{IEEEbiography}

\vspace{-1cm}

\begin{IEEEbiography}[{\includegraphics[width=1in,height=1.25in,clip,keepaspectratio]{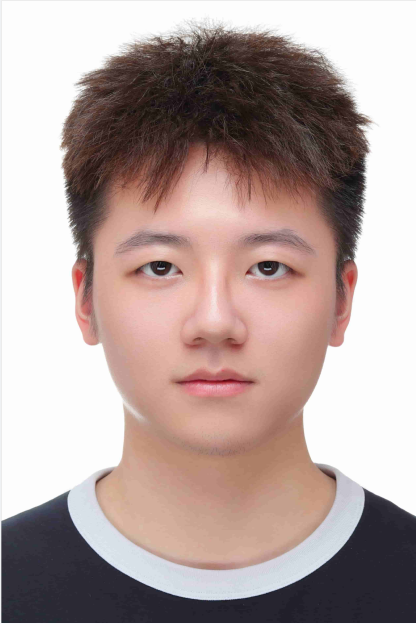}}]{Haibo Zhao} received the B.S degree from Tongji University, Shanghai, China, in 2024. He is currently pursuing the Master of Science in big data technology from the Department of Computer Science and Engineering, Hong Kong University of Science and Technology, Hong Kong, China. His current research interests include distributed learning, vision language model and recommender system.
\end{IEEEbiography}

% \vspace{-1.5cm}

\section{Appendix}
% \vspace{-1cm}
This appendix provides the main proof process for Theorem 1. More details can be found in the file\footnote{\url{https://anonymous.4open.science/r/Argus_TNSE}}.

% \vspace{-2cm}

\subsection{Proof of Theorem 1}
% \vspace{-0.5cm}
% Lemma 1
\subsubsection{Lemma 1 (Descending Inequality of x Variables)} 
Under Assumptions 1 and 2, by applying the Cauchy-Schwarz inequality and Jensen's inequality, we have:
% \vspace{-1cm}
\begin{footnotesize}
\begin{equation}
\begin{aligned}
& \mathbb{E}[L_{p}(\{\bar{\mathbf{x}}^{t+1}\},\{\mathbf{y}^{t}_i\}, \{\lambda_{i,l}^{t}\},\{\boldsymbol{\theta}_{i,j}^{t}\}) - L_{p}(\{\bar{\mathbf{x}}^{t}\},\{\mathbf{y}^{t}_i\}, \{\lambda_{i,l}^{t}\},\{\boldsymbol{\theta}_{i,j}^{t}\})]\\
\le & (NL- \frac{N}{2\eta_x} )\mathbb{E}[\sum_{i=1}^N ||\mathbf{x}_i^{t+1} -\bar{\mathbf{x}}^{t}||^2]   - \frac{N}{2\eta_x} \mathbb{E}[\sum_{i=1}^N||\mathbf{x}_i^{t+1}-\mathbf{d}_i^{t}||^2] \\
+ & (\frac{N}{2\eta_x}+N^2L) \mathbb{E}[\sum_{i=1}^N||\bar{\mathbf{x}}^{t}-\mathbf{x}_i^{t}||^2]\\
+ & 2\tau k_1N^2L(\mathbb{E}[\sum_{i=1}^N(||\mathbf{x}_i^{t+1} - \mathbf{x}_i^{t}||^2] + \mathbb{E}[\sum_{i=1}^N||\mathbf{y}_i^{t+1} - \mathbf{y}_i^{t}||^2])),
\end{aligned}
\label{asy_eq_lemma1_1}
\end{equation}
\end{footnotesize}
where $\bar{\mathbf{x}}^{t+1} = \frac{1}{N}\sum^{N}_{i=1}\mathbf{x}_i^{t+1}$, $\bar{\mathbf{x}}^t = \frac{1}{N}\sum^{N}_{i=1}\mathbf{x}_i^t$.

% Lemma 2
\subsubsection{Lemma 2 (Descending Inequality of y Variables)} 
\label{lemma2}
Under Assumptions 1 and 2, by applying the Cauchy-Schwarz inequality and the triangle inequality, the following inequality holds:
% \vspace{-0.5cm}
\begin{footnotesize}
\begin{equation}
\begin{aligned}
& \mathbb{E}[L_{p}(\{\bar{\mathbf{x}}^{t+1}\},\{\bar{\mathbf{y}}^{t+1}\}, \{\lambda_{i,l}^{t}\},\{\boldsymbol{\theta}_{i,j}^{t}\}) - L_{p}(\{\bar{\mathbf{x}}^{t+1}\},\{\bar{\mathbf{y}}^{t}\}, \{\lambda_{i,l}^{t}\},\{\boldsymbol{\theta}_{i,j}^{t}\})]\\
\le &  \frac{3\eta_yN^2L^2}{2\beta}(\mathbb{E}[\sum_{i=1}^N||\mathbf{x}^{t+1} -\bar{\mathbf{x}}^{t}||^2] + \mathbb{E}[\sum_{i=1}^N||\bar{\mathbf{x}}^{t} -\mathbf{x}_i^{t}||^2] + \mathbb{E}[\sum_{i=1}^N||\bar{\mathbf{y}}^{t} -\mathbf{y}_i^{t}||^2]) \\
+ & \frac{3\eta_yN^2L^2\tau k_1}{\beta}(\mathbb{E}[\sum_{i=1}^N||\mathbf{x}_i^{t+1} -\mathbf{x}_i^{t}||^2] + \mathbb{E}[\sum_{i=1}^N||\mathbf{y}_i^{t+1} -\mathbf{y}_i^{t}||^2)\\
+& (\frac{\eta_yN\beta}{2}+\frac{\eta_y^2NL}{2}-\eta_y N) \mathbb{E}[\sum_{i=1}^{N}||\nabla_{\mathbf{y}_i} \tilde{L}_{pi}(\{\mathbf{x}_i^{\hat{t}_i}\}, \{\mathbf{y}_i^{\hat{t}_i}\}, \{\lambda_{i,l}^{\hat{t}_i} \},\{\boldsymbol{\theta}^{\hat{t}_i}_{i,j}\})||^2],
\end{aligned}
\label{asy_eq_lemma2}
\end{equation}
\end{footnotesize}
\noindent where $\bar{\mathbf{y}}^{t+1} = \frac{1}{N}\sum^{N}_{i=1}\mathbf{y}_i^{t+1}$, $\bar{\mathbf{y}}^t = \frac{1}{N}\sum^{N}_{i=1}\mathbf{y}_i^t$, $\beta$ is a parameter that can be tuned later.

% Lemma 3
\subsubsection{Lemma 3 (Descending Inequality of  the $L_p$ Function)} 
\label{lemma3}
Under Assumptions 1 and 2, based on Lemmas 1 and 2, and by applying the Cauchy-Schwarz inequality and the triangle inequality, the following inequality holds: 

\begin{footnotesize}
\begin{equation*}
\begin{aligned}
& \mathbb{E}[L_{p}(\{\bar{\mathbf{x}}^{t+1}\},\{\bar{\mathbf{y}}^{t+1}\}, \{\lambda_{i,l}^{t+1}\},\{\boldsymbol{\theta}_{i,j}^{t+1}\}) - L_{p}(\{\bar{\mathbf{x}}^{t}\},\{\bar{\mathbf{y}}^{t}\}, \{\lambda_{i,l}^{t}\},\{\boldsymbol{\theta}_{i,j}^{t}\})]\\
\le & (NL- \frac{N}{2\eta_x} + \frac{3\eta_yN^2L^2}{2\beta} + \frac{MNL^2\eta_{\lambda}}{2} + \frac{N^2L^2\eta_{\boldsymbol{\theta}}}{2})\mathbb{E}[\sum_{i=1}^N ||\mathbf{x}_i^{t+1} -\bar{\mathbf{x}}^{t}||^2]   \\
- & \frac{N}{2\eta_x} \mathbb{E}[\sum_{i=1}^N||\mathbf{x}_i^{t+1}-\mathbf{d}_i^{t}||^2] \\
+& (\frac{N}{2\eta_x}+N^2L +  \frac{3\eta_yN^2L^2}{2\beta} + \frac{MNL^2\eta_{\lambda}}{2} + \frac{N^2L^2\eta_{\boldsymbol{\theta}}}{2}) \mathbb{E}[\sum_{i=1}^N||\bar{\mathbf{x}}^{t}-\mathbf{x}_i^{t}||^2]\\
+ & (\frac{3\eta_yN^2L^2}{2\beta} + \frac{MNL^2\eta_{\lambda}}{2} + \frac{N^2L^2\eta_{\boldsymbol{\theta}}}{2})
(\mathbb{E}[\sum_{i=1}^N||\bar{\mathbf{y}}^{t} -\mathbf{y}_i^{t}||^2])\\
+ & (2\tau k_1N^2L + \frac{3\eta_yN^2L^2\tau k_1}{\beta})(\mathbb{E}[\sum_{i=1}^N(||\mathbf{x}_i^{t+1} - \mathbf{x}_i^{t}||^2] + \mathbb{E}[\sum_{i=1}^N||\mathbf{y}_i^{t+1} - \mathbf{y}_i^{t}||^2]))\\
+&(\frac{\eta_yN\beta}{2}+\frac{\eta_y^2NL}{2}-\eta_y N + \frac{MNL^2\eta_y^2\eta_{\lambda}}{2} + \frac{N^2L^2\eta_y^2\eta_{\boldsymbol{\theta}}}{2}) \\
& \mathbb{E}[\sum_{i=1}^{N}||\nabla_{\mathbf{y}_i} \tilde{L}_{pi}(\{\mathbf{x}_i^{\hat{t}_i}\}, \{\mathbf{y}_i^{\hat{t}_i}\}, \{\lambda_{i,l}^{\hat{t}_i} \},\{\boldsymbol{\theta}^{\hat{t}_i}_{i,j}\})||^2]\\
+ & (\frac{1}{\eta_{\lambda}}-\frac{c_{1}^{t-1}-c_{1}^{t}}{2})\sum_{i=1}^{N}\sum_{l=1}^{|\mathcal{P}_i^t|}\mathbb{E}[||\lambda_{i,l}^{t+1}-\lambda_{i,l}^{t}||^2]  \\
+ & (\frac{1}{\eta_{\theta}}-\frac{c_{2}^{t-1}-c_{2}^{t}}{2})\mathbb{E}[\sum_{i=1}^{N}\sum_{j=1}^{N}||\boldsymbol{\theta}_{i,j}^{t+1}-\boldsymbol{\theta}_{i,j}^{t}||^2] \\
\end{aligned}
\label{asy_eq_lemma3_1}
\end{equation*}
\end{footnotesize}

% tem
\begin{footnotesize}
\begin{equation}
\begin{aligned}
+ & \frac{c_1^{t-1}}{2}\mathbb{E}[\sum_{i=1}^{N}\sum_{l=1}^{|\mathcal{P}^{t}_i|}(||\lambda_{i,l}^{t+1}||^2 - ||\lambda_{i,l}^{t}||^2)] + \frac{1}{2 \eta_{\lambda}}\mathbb{E}[\sum_{i=1}^{N}\sum_{l=1}^{|\mathcal{P}_i^t|}||\lambda_{i,l}^{t}-\lambda_{i,l}^{t-1}||^{2}]\\
+ & \frac{c_2^{t-1}}{2}\mathbb{E}[\sum_{i=1}^{N}\sum_{j=1}^{N}(||\boldsymbol{\theta}_{i,j}^{t+1}||^2 - ||\boldsymbol{\theta}_{i,j}^{t}||^2]) + \frac{1}{2 \eta_{\theta}}\sum_{i=1}^{N}\mathbb{E}[\sum_{j=1}^{N}||\boldsymbol{\theta}_{i,j}^{t}-\boldsymbol{\theta}_{i,j}^{t-1}||^{2}].
\end{aligned}
\label{asy_eq_lemma3}
\end{equation}
\end{footnotesize}

% Lemma 4
\subsubsection{Lemma 4 (Iterates Contraction)} 
\label{lemma4}
Under Assumptions 1 and 2, by applying the Peter-Paul inequality, the following contraction properties of iterates hold:
\begin{footnotesize}
\begin{equation}
\begin{aligned}
& \mathbb{E}[\sum_{i=1}^N||\mathbf{x}_i^{t+1} - \bar{\mathbf{x}}^{t+1}||^{2}] {\le}(\rho + \frac{170\eta_x^2 L^2}{1-\rho})\mathbb{E}[\sum_{i=1}^N||\mathbf{x}_i^{t} - \bar{\mathbf{x}}^{t}||^{2}] \\
&+ \frac{40\eta_x^2 L^2}{1-\rho}\mathbb{E}[\sum_{i=1}^N||\mathbf{x}_i^{t+1} - \mathbf{d}_i^{t}||^2] + \frac{60\eta_x^2L^2}{(1-\rho)^2}\mathbb{E}[\sum_{i=1}^N||\bar{\mathbf{y}}^{t} - \mathbf{y}_i^{t}||^2] \\
& + \frac{40\eta_x^2\eta_y^2 L^2}{(1-\rho)^2}\mathbb{E}[\sum_{i=1}^N||\nabla_{\mathbf{y}_i} \tilde{L}_{pi}(\{\mathbf{x}_i^{\hat{t}_i}\}, \{\mathbf{y}_i^{\hat{t}_i}\}, \{\lambda_{i,l}^{\hat{t}_i} \},\{\boldsymbol{\theta}^{\hat{t}_i}_{i,j}\})||^2],
\end{aligned}
\label{asy_eq_lemma4_1}
\end{equation}
\end{footnotesize}

\begin{footnotesize}
\begin{equation}
\begin{aligned}
&\mathbb{E}[\sum_{i=1}^N||\mathbf{y}_i^{t+1} - \bar{\mathbf{y}}^{t+1}||^{2}] {\le} (\rho + \frac{230\eta_y^2 L^2}{(1-\rho)^2})\mathbb{E}[\sum_{i=1}^N||\mathbf{y}_i^{t} - \bar{\mathbf{y}}^{t}||^{2}] \\
& + \frac{40\eta_y^2 L^2}{1-\rho}\mathbb{E}[\sum_{i=1}^N||\mathbf{x}_i^{t+1} - \mathbf{d}_i^{t}||^2] \\
& + \frac{40\eta_y^4 L^2}{(1-\rho)^2}\mathbb{E}[\sum_{i=1}^N||\nabla_{\mathbf{y}_i} \tilde{L}_{pi}(\{\mathbf{x}_i^{\hat{t}_i}\}, \{\mathbf{y}_i^{\hat{t}_i}\}, \{\lambda_{i,l}^{\hat{t}_i} \},\{\boldsymbol{\theta}^{\hat{t}_i}_{i,j}\})||^2].
\end{aligned}
\label{asy_eq_lemma4_2}
\end{equation}
\end{footnotesize}

% Lemma 5
\subsubsection{Lemma 5}
\label{lemma5} Denoting $S_1^{t+1}$, $S_2^{t+1}$, and $F^{t+1}$ as,

\begin{footnotesize}
\begin{equation}
\begin{aligned}
S_{1}^{t+1} & =\frac{4}{\eta_{\lambda}^{2} c_{1}^{t+1}} \sum_{i=1}^N\sum_{l=1}^{\left|\mathcal{P}_i^{t}\right|}||\lambda_{i,l}^{t+1}-\lambda_{i,l}^{t}||^{2}\\
&-\frac{4}{\eta_{\lambda}}(\frac{c_{1}^{t-1}}{c_{1}^{t}}-1) \sum_{i=1}^N\sum_{l=1}^{\left|\mathcal{P}_i^{t}\right|}||\lambda_{i,l}^{t+1}||^{2},
\end{aligned}
\label{asy_eq_lemma5_1}
\end{equation}
\end{footnotesize}

\begin{footnotesize}
\begin{equation}
\begin{aligned}
S_{2}^{t+1} & =\frac{4}{\eta_{\theta}^{2} c_{2}^{t+1}} \sum_{i=1}^{N}\sum_{j=1}^{N}||\boldsymbol{\theta}_{i,j}^{t+1}-\boldsymbol{\theta}_{i,j}^{t}||^{2}\\
&-\frac{4}{\eta_{\theta}}(\frac{c_{2}^{t-1}}{c_{2}^{t}}-1) \sum_{i=1}^{N}\sum_{j=1}^{N}||\boldsymbol{\theta}_{i,j}^{t+1}||^{2},
\end{aligned}
\label{asy_eq_lemma5_2}
\end{equation}
\end{footnotesize}

\begin{footnotesize}
\begin{equation}
\begin{aligned}
&F^{t+1}=L_{p}(\{\bar{\mathbf{x}}^{t+1}\},\{\bar{\mathbf{y}}^{t+1}\}, \{\lambda_{i,l}^{t+1}\},\{\boldsymbol{\theta}_{i,j}^{t+1}\}) + S_1^{t+1} + S_2^{t+1} \\
&+ \gamma_1^{t}\sum_{i=1}^{N}||\mathbf{x}_i^{t+1} - \bar{\mathbf{x}}^{t+1}||^2+ \gamma_2^{t} \sum_{i=1}^{N}||\mathbf{y}_i^{t+1} - \bar{\mathbf{y}}^{t+1}||^2\\
& - \frac{6}{\eta_{\lambda}} \sum_{i=1}^{N}\sum_{l=1}^{\left|\mathcal{P}_i^{t}\right|}||\lambda_{i,l}^{t+1}-\lambda_{i,l}^{t}||^{2}-\frac{c_{1}^{t}}{2} \sum_{i=1}^{N}\sum_{l=1}^{\left|\mathcal{P}_i^{t}\right|}||\lambda_{i,l}^{t+1}||^{2} \\
& -\frac{6}{\eta_{\theta}} \sum_{i=1}^{N}\sum_{j=1}^{N}||\boldsymbol{\theta}_{i,j}^{t+1}-\boldsymbol{\theta}_{i,j}^{t}||^{2}-\frac{c_{2}^{t}}{2} \sum_{i=1}^{N}\sum_{j=1}^{N}||\boldsymbol{\theta}_{i,j}^{t+1}||^{2},
\end{aligned}
\label{asy_eq_lemma5_3}
\end{equation}
\end{footnotesize}

\hspace{1em} $\forall t \ge T_1$, we have

\begin{footnotesize}
\begin{equation*}
\begin{aligned}
& (\frac{N}{2{\eta_x}} - NL - \frac{3\eta_yN^2L^2}{2{\beta}} - \frac{MNL^2{\eta_{\lambda}}}{2} - \frac{N^2L^2{\eta_{\boldsymbol{\theta}}}}{2})\mathbb{E}[\sum_{i=1}^N ||\mathbf{x}_i^{t+1} -\bar{\mathbf{x}}^{t}||^2]  \\
&+ (\frac{N}{2{\eta_x}} - (4\tau k_1N^2L + \frac{8\eta_yN^2L^2\tau k_1}{{\beta}} + \frac{16NML^2}{{\eta_{\lambda}}(c_1^t)^2 } + \frac{16N^2L^2}{\eta_{\theta}(c_2^t)^2}) \\
&- \frac{40L^2(\eta_x^2\gamma_1^t + \eta_y^2\gamma_2^t)}{1-\rho}) \mathbb{E}[\sum_{i=1}^N||\mathbf{x}_i^{t+1}-\mathbf{d}_i^{t}||^2] \\
& + (\gamma_1^{t-1} - \gamma_1^{t}(\rho + \frac{170\eta_x^2L^2}{1-\rho})- \frac{N}{2{\eta_x}} - N^2L -  \frac{3\eta_yN^2L^2}{2{\beta}} - \frac{MNL^2{\eta_{\lambda}}}{2} \\
& - \frac{N^2L^2{\eta_{\boldsymbol{\theta}}}}{2} - (16\tau k_1N^2L + \frac{32\eta_yN^2L^2\tau k_1}{{\beta}} + \frac{64NML^2}{{\eta_{\lambda}}(c_1^t)^2 } + \frac{64N^2L^2}{\eta_{\theta}(c_2^t)^2}))\\
\end{aligned}
\label{eq_lemma5_4_1}
\end{equation*}
\end{footnotesize}

\begin{footnotesize}
\begin{equation}
\begin{aligned}
& \mathbb{E}[\sum_{i=1}^N||\bar{\mathbf{x}}^{t}-\mathbf{x}_i^{t}||^2] + (\gamma_2^{t-1} - \gamma_2^{t}(\rho+\frac{230\eta_y^2L^2}{1-\rho}) - \frac{3\eta_yN^2L^2}{2{\beta}} \\
& - \frac{MNL^2{\eta_{\lambda}}}{2} - \frac{N^2L^2{\eta_{\boldsymbol{\theta}}}}{2} - \frac{60\eta_x^2L^2\gamma_1^t}{1-\rho} - (\rho + \frac{2}{1-\rho})(2\tau k_1N^2L \\
& + \frac{4\eta_yN^2L^2\tau k_1}{{\beta}} + \frac{8NML^2}{{\eta_{\lambda}}(c_1^t)^2 } + \frac{8N^2L^2}{\eta_{\theta}(c_2^t)^2}))
\mathbb{E}[\sum_{i=1}^N||\bar{\mathbf{y}}^{t} -\mathbf{y}_i^{t}||^2] \\
&+(\eta_y N - \frac{\eta_yN{\beta}}{2} - \frac{\eta_y^2NL}{2} - \frac{MNL^2\eta_y^2{\eta_{\lambda}}}{2} - \frac{N^2L^2\eta_y^2{\eta_{\boldsymbol{\theta}}}}{2} \\
&- \frac{40\eta_y^2 L^2 (\eta_x^2\gamma_1^t+\eta_y^2\gamma_2^t)}{(1-\rho)^2} - \frac{2\eta_y^2}{1-\rho}(2\tau k_1N^2L + \frac{4\eta_yN^2L^2\tau k_1}{{\beta}} \\
& + \frac{8NML^2}{{\eta_{\lambda}}(c_1^t)^2 } + \frac{8N^2L^2}{\eta_{\theta}(c_2^t)^2})) \mathbb{E}[\sum_{i=1}^{N}||\nabla_{\mathbf{y}_i} \tilde{L}_{pi}(\{\mathbf{x}_i^{\hat{t}_i}\}, \{\mathbf{y}_i^{\hat{t}_i}\}, \{\lambda_{i,l}^{\hat{t}_i} \},\{\boldsymbol{\theta}^{\hat{t}_i}_{i,j}\})||^2]\\
&+ \frac{1}{{\eta_{\lambda}}}\mathbb{E}[\sum_{i=1}^{N}\sum_{l=1}^{|\mathcal{P}_i^t|}||\lambda_{i,l}^{t+1}-\lambda_{i,l}^{t}||^2]  + \frac{1}{\eta_{\theta}}\mathbb{E}[\sum_{i=1}^{N}\sum_{j=1}^{N}||\boldsymbol{\theta}_{i,j}^{t+1}-\boldsymbol{\theta}_{i,j}^{t}||^2] \\
&\le \mathbb{E}[F^t-F^{t+1}] + \frac{4}{{\eta_{\lambda}}}(\frac{c_{1}^{t-2}}{c_{1}^{t-1}}-\frac{c_{1}^{t-1}}{c_{1}^{t}}) \mathbb{E}[\sum_{i=1}^N\sum_{l=1}^{|\mathcal{P}_i^{t}|}||\lambda_{i,l}^{t}||^{2}] \\
& +  \frac{4}{\eta_{\theta}}(\frac{c_{2}^{t-2}}{c_{2}^{t-1}}-\frac{c_{2}^{t-1}}{c_{2}^{t}}) \mathbb{E}[\sum_{i=1}^N\sum_{j=1}^N||\boldsymbol{\theta}_{i,j}^{t}||^{2}]\\
&+ (\frac{c_1^{t-1}}{2} - \frac{c_1^{t}}{2})\mathbb{E}[\sum_{i=1}^{N}\sum_{l=1}^{|\mathcal{P}^{t}_i|}||\lambda_{i,l}^{t+1}||^2 ] + (\frac{c_2^{t-1}}{2} - \frac{c_2^{t}}{2})\mathbb{E}[\sum_{i=1}^{N}\sum_{j=1}^{N}||\boldsymbol{\theta}_{i,j}^{t+1}||^2].
\end{aligned}
\label{eq_lemma5_4}
\end{equation}
\end{footnotesize}

% 简写一下推导过程
\subsubsection{Proof of Theorem 1}
\begin{proof}
According to Assumption 1, the property of the proximal operator, and Young’s inequality, we have:

\begin{footnotesize}
\begin{equation}
\begin{aligned}
& \mathbb{E}[\sum_{i=1}^N||P(\mathbf{d}_i^t, \bar{\nabla}_{\mathbf{d}} L'_{pi}(\left\{\mathbf{d}_i^t\right\},\left\{\mathbf{y}_{i}^{t}\right\}, \left\{\lambda_{i,l}^{t}\right\},\{\boldsymbol{\theta}_{i,j}^{t}\}), \eta_{i,x})||^2]\\
& + L^2\mathbb{E}[\sum_{i=1}^{N}||\mathbf{x}_i^t - \bar{\mathbf{x}}^t ||^2]
\le ({\frac{2}{\eta_{x}^2} + 32NL^2\tau k_1) \mathbb{E}[\sum_{i=1}^N||\mathbf{x}_i^{t+1} - \mathbf{d}_i^{t}||^2]} \\
& + (128NL^2\tau k_1+32NL^2) \mathbb{E}[\sum_{i=1}^N||\mathbf{x}_i^{t} - \bar{\mathbf{x}}^{t}||^2] +\frac{48NL^2\tau k_1}{1-\rho}\mathbb{E}[\sum_{i=1}^N||\mathbf{y}_i^{t} - \bar{\mathbf{y}}^{t}||^2] \\
&+ \frac{32\eta_y^2NL^2\tau k_1}{1-\rho}\mathbb{E}[\sum_{i=1}^N||\nabla_{\mathbf{y}_i} \tilde{L}_{pi}(\{\mathbf{x}_i^{\hat{t}_i}\}, \{\mathbf{y}_i^{\hat{t}_i}\}, \{\lambda_{i,l}^{\hat{t}_i} \},\{\boldsymbol{\theta}^{\hat{t}_i}_{i,j}\})||^2],
\end{aligned}
\label{asy_theorem1_10}
\end{equation}
\end{footnotesize}

and 

\begin{footnotesize}
\begin{equation}
\begin{aligned}
& \frac{1}{2}\mathbb{E}[\sum_{i=1}^N||\bar{\nabla}_{\mathbf{u}} L_{pi}'(\{\mathbf{x}_i^t\},\{\mathbf{u}_{i}^{t}\}, \{\lambda_{i,l}^{t}\},\{\boldsymbol{\theta}_{i,j}^{t}\})||^2] \\
& \le \mathbb{E}[\sum_{i=1}^N||\nabla_{\mathbf{y}_i} \tilde{L}_{pi}(\{\mathbf{x}_i^{\hat{t}_i}\}, \{\mathbf{y}_i^{\hat{t}_i}\}, \{\lambda_{i,l}^{\hat{t}_i} \},\{\boldsymbol{\theta}^{\hat{t}_i}_{i,j}\})||^2]\\
& + \mathbb{E}[\sum_{i=1}^N||\nabla_{\mathbf{y}_i}\tilde{L}_{pi}(\{\mathbf{x}_i^{\hat{t}_i}\}, \{\mathbf{y}_i^{\hat{t}_i}\}, \{\lambda_{i,l}^{\hat{t}_i} \},\{\boldsymbol{\theta}^{\hat{t}_i}_{i,j}\}) \\
& - \bar{\nabla}_{\mathbf{u}} L_{pi}'(\{\mathbf{x}_i^t\},\{\mathbf{u}_{i}^{t}\}, \{\lambda_{i,l}^{t}\},\{\boldsymbol{\theta}_{i,j}^{t}\})||^2].
\end{aligned}
\label{asy_theorem1_12}
\end{equation}
\end{footnotesize}

Given the definition of $(\tilde{{\mathcal{G}}}^{t})_{\lambda_{i,l}} = \nabla_{\lambda_{i,l}}\tilde{L}_{p}(\{\mathbf{x}_i^{t}\},\{\mathbf{y}^{t}_i\}, $

\noindent $\{\lambda_{i,l}^{t}\}, \{\boldsymbol{\theta}_{i,j}^{t}\})$ and 
$(\tilde{{\mathcal{G}}}^{t})_{\boldsymbol{\theta}_{i,j}} = \nabla_{\boldsymbol{\theta}_{i,j}}\tilde{L}_{p}(\{\mathbf{x}_i^{t}\},\{\mathbf{y}^{t}_i\},  \{\lambda_{i,l}^{t}\},$

\noindent $ \{\boldsymbol{\theta}_{i,j}^{t}\})$, according to the update rules,  trigonometric inequality, and Cauchy-Schwarz inequality, we have:

\begin{footnotesize}
\begin{equation*}
\begin{aligned}
& \mathbb{E}[||(\tilde{\mathcal{G}}^{t})_{\lambda_{i,l}}||^2] 
\le 3L^2(\mathbb{E}[\sum_{i=1}^N||\mathbf{x}_i^{t+1} -\mathbf{x}_i^{t}||^2] + \mathbb{E}[\sum_{i=1}^N||\mathbf{y}_i^{t+1} -\mathbf{y}_i^{t}||^2]) \\
\end{aligned}
\label{asy_theorem1_18_1}
\end{equation*}
\end{footnotesize}

% tem
\begin{footnotesize}
\begin{equation}
\begin{aligned}
& + 3((c_1^{t-1})^2 - (c_1^{t})^2)\mathbb{E}[||\lambda_{i,l}^{t}||^2] + \frac{3}{\eta_{\lambda}^2}\mathbb{E}[||\lambda_{i,l}^{t+1}-\lambda_{i,l}^{t}||^2]\\
& \le 6L^2\mathbb{E}[\sum_{i=1}^N||\mathbf{x}_i^{t+1} - \mathbf{d}_i^{t}||^2] + 24L^2 \mathbb{E}[\sum_{i=1}^N||\mathbf{x}_i^{t} - \bar{\mathbf{x}}^{t}||^2] \\
& + \frac{9L^2}{1-\rho}\mathbb{E}[\sum_{i=1}^N||\mathbf{y}_i^{t} - \bar{\mathbf{y}}^{t}||^2] \\
& + \frac{6L^2\eta_y^2}{1-\rho}\mathbb{E}[\sum_{i=1}^N||\nabla_{\mathbf{y}_i} \tilde{L}_{pi}(\{\mathbf{x}_i^{\hat{t}_i}\}, \{\mathbf{y}_i^{\hat{t}_i}\}, \{\lambda_{i,l}^{\hat{t}_i} \},\{\boldsymbol{\theta}^{\hat{t}_i}_{i,j}\})||^2] \\
& + 3((c_1^{t-1})^2 - (c_1^{t})^2)\mathbb{E}[||\lambda_{i,l}^{t}||^2] + \frac{3}{\eta_{\lambda}^2}\mathbb{E}[||\lambda_{i,l}^{t+1}-\lambda_{i,l}^{t}||^2],
\end{aligned}
\label{asy_theorem1_18}
\end{equation}
\end{footnotesize}

and 

\begin{footnotesize}
\begin{equation}
\begin{aligned}
& \mathbb{E}[||(\tilde{\mathcal{G}}^{t})_{\boldsymbol{\theta}_{i,j}}||^2]
\le 3L^2(\mathbb{E}[\sum_{i=1}^N||\mathbf{x}_i^{t+1} -\mathbf{x}_i^{t}||^2] + \mathbb{E}[\sum_{i=1}^N||\mathbf{y}_i^{t+1} -\mathbf{y}_i^{t}||^2]) \\
& + 3((c_2^{t-1})^2 - (c_2^{t})^2)\mathbb{E}[||\boldsymbol{\theta}_{i,j}^{t}||^2] + \frac{3}{\eta_{\theta}^2}\mathbb{E}[||\boldsymbol{\theta}_{i,j}^{t+1}-\boldsymbol{\theta}_{i,j}^{t}||^2]\\
& \le 6L^2\mathbb{E}[\sum_{i=1}^N||\mathbf{x}_i^{t+1} - \mathbf{d}_i^{t}||^2] + 24L^2 \mathbb{E}[\sum_{i=1}^N||\mathbf{x}_i^{t} - \bar{\mathbf{x}}^{t}||^2] \\
& + \frac{9L^2}{1-\rho}\mathbb{E}[\sum_{i=1}^N||\mathbf{y}_i^{t} - \bar{\mathbf{y}}^{t}||^2] \\
& + \frac{6L^2\eta_y^2}{1-\rho}\mathbb{E}[\sum_{i=1}^N||\nabla_{\mathbf{y}_i} \tilde{L}_{pi}(\{\mathbf{x}_i^{\hat{t}_i}\}, \{\mathbf{y}_i^{\hat{t}_i}\}, \{\lambda_{i,l}^{\hat{t}_i} \},\{\boldsymbol{\theta}^{\hat{t}_i}_{i,j}\})||^2] \\
& + 3((c_2^{t-1})^2 - (c_2^{t})^2)\mathbb{E}[||\boldsymbol{\theta}_{i,j}^{t}||^2] + \frac{3}{\eta_{\theta}^2}\mathbb{E}[||\boldsymbol{\theta}_{i,j}^{t+1}-\boldsymbol{\theta}_{i,j}^{t}||^2].
\end{aligned}
\label{asy_theorem1_21}
\end{equation}
\end{footnotesize}

According to the inequality of norms squared differences, we have:

\begin{equation}
\begin{aligned}
& \mathbb{E}[\Psi^t] - \mathbb{E}[\tilde{\Psi}^t] \\
&= \sum_{i=1}^N\sum_{l=1}^{|\mathcal{P}_i|}(\mathbb{E}[||({\mathcal{G}}^{t})_{\lambda_{i,l}}||^2] - \mathbb{E}[||(\tilde{{\mathcal{G}}}^{t})_{\lambda_{i,l}}||^2]) \\
&+ \sum_{i=1}^N\sum_{j=1}^{N}(\mathbb{E}[||({\mathcal{G}}^{t})_{\boldsymbol{\theta}_{i,j}}||^2] - \mathbb{E}[||(\tilde{{\mathcal{G}}}^{t})_{\boldsymbol{\theta}_{i,j}}||^2])\\
\le & \sum_{i=1}^N\sum_{l=1}^{|\mathcal{P}_i^t|}\mathbb{E}[||c_1^{t-1} \lambda_{i,l}^t||^2] +  \sum_{i=1}^N\sum_{j=1}^{N}\mathbb{E}[||c_2^{t-1} \boldsymbol{\theta}_{i,j}^t||^2].
\end{aligned}
\label{asy_theorem1_39}
\end{equation}

According to Lemma4, Eq.(\ref{asy_theorem1_10}), Eq.(\ref{asy_theorem1_12}), Eq.(\ref{asy_theorem1_18}) Eq.(\ref{asy_theorem1_21}), and Eq.(\ref{asy_theorem1_39}), we can obtain:

\begin{footnotesize}
\begin{equation*}
\begin{aligned}
&\mathbb{E}[\Psi^t]\le  C_1' \mathbb{E}[\sum_{i=1}^N ||\mathbf{x}_i^{t+1} -\bar{\mathbf{x}}^{t}||^2] + C_2' \mathbb{E}[\sum_{i=1}^N||\mathbf{x}_i^{t+1} - \mathbf{d}_i^{t}||^2] \\
&+  C_3' \mathbb{E}[\sum_{i=1}^{N}||\mathbf{x}_i^t - \bar{\mathbf{x}}^t ||^2] + C_4'(\mathbb{E}[\sum_{i=1}^N||\bar{\mathbf{y}}^{t}-\mathbf{y}_i^t||^2]) \\
&+  C_5' \mathbb{E}[\sum_{i=1}^N||\nabla_{\mathbf{y}_i} \tilde{L}_{pi}(\{\mathbf{x}_i^{\hat{t}_i}\}, \{\mathbf{y}_i^{\hat{t}_i}\}, \{\lambda_{i,l}^{\hat{t}_i} \},\{\boldsymbol{\theta}^{\hat{t}_i}_{i,j}\})||^2] 
\\
&+ C_6' \sum_{i=1}^{N}\sum_{l=1}^{|\mathcal{P}^{t}_i|}\mathbb{E}[||\lambda_{i,l}^{t+1}-\lambda_{i,l}^{t}||^2] + C_7'\sum_{i=1}^N\sum_{j=1}^N\mathbb{E}[||\boldsymbol{\theta}_{i,j}^{t+1}-\boldsymbol{\theta}_{i,j}^{t}||^2]  \\
&+ (4(c_1^{t-1})^2 - 3(c_1^{t})^2)\sum_{i=1}^{N}\sum_{l=1}^{|\mathcal{P}^{t}_i|}\mathbb{E}[||\lambda_{i,l}^{t}||^2] \\
\end{aligned}
\label{asy_theorem1_26_1}
\end{equation*}
\end{footnotesize}

% tem
\begin{footnotesize}
\begin{equation}
\begin{aligned}
&+ 4((c_2^{t-1})^2 -3(c_2^{t})^2)\sum_{i=1}^N\sum_{j=1}^N\mathbb{E}[||\boldsymbol{\theta}_{i,j}^{t}||^2].
\end{aligned}
\label{asy_theorem1_26}
\end{equation}
\end{footnotesize}

According to Lemma5, let $\eta_x = \eta_y = \eta_{\lambda} = \eta_{\theta} = \eta$, $\beta = 1$ to have:

\begin{footnotesize}
\begin{equation}
\begin{aligned}
& C_1 \mathbb{E}[\sum_{i=1}^N ||\mathbf{x}_i^{t+1} -\bar{\mathbf{x}}^{t}||^2]  + C_2 \mathbb{E}[\sum_{i=1}^N||\mathbf{x}_i^{t+1}-\mathbf{d}_i^{t}||^2] \\
& + C_3 \mathbb{E}[\sum_{i=1}^N||\bar{\mathbf{x}}^{t}-\mathbf{x}_i^{t}||^2]
+ C_4 \mathbb{E}[\sum_{i=1}^N||\bar{\mathbf{y}}^{t} -\mathbf{y}_i^{t}||^2] \\
& + C_5 \mathbb{E}[\sum_{i=1}^{N}||\nabla_{\mathbf{y}_i} \tilde{L}_{pi}(\{\mathbf{x}_i^{\hat{t}_i}\}, \{\mathbf{y}_i^{\hat{t}_i}\}, \{\lambda_{i,l}^{\hat{t}_i} \},\{\boldsymbol{\theta}^{\hat{t}_i}_{i,j}\})||^2] \\
& +  C_6 \mathbb{E}[\sum_{i=1}^{N}\sum_{l=1}^{|\mathcal{P}_i^t|}||\lambda_{i,l}^{t+1}-\lambda_{i,l}^{t}||^2]  + C_7 \mathbb{E}[\sum_{i=1}^{N}\sum_{j=1}^{N}||\boldsymbol{\theta}_{i,j}^{t+1}-\boldsymbol{\theta}_{i,j}^{t}||^2] \\
& \le  \mathbb{E}[F^t-F^{t+1}] + \frac{4}{{\eta}}(\frac{c_{1}^{t-2}}{c_{1}^{t-1}}-\frac{c_{1}^{t-1}}{c_{1}^{t}}) \mathbb{E}[\sum_{i=1}^N\sum_{l=1}^{|\mathcal{P}_i^{t}|}||\lambda_{i,l}^{t}||^{2}] \\
& +  \frac{4}{\eta}(\frac{c_{2}^{t-2}}{c_{2}^{t-1}}-\frac{c_{2}^{t-1}}{c_{2}^{t}}) \mathbb{E}[\sum_{i=1}^N\sum_{j=1}^N||\boldsymbol{\theta}_{i,j}^{t}||^{2}]\\
& + (\frac{c_1^{t-1}}{2} - \frac{c_1^{t}}{2})\mathbb{E}[\sum_{i=1}^{N}\sum_{l=1}^{|\mathcal{P}^{t}_i|}||\lambda_{i,l}^{t+1}||^2 ] + (\frac{c_2^{t-1}}{2} - \frac{c_2^{t}}{2})\mathbb{E}[\sum_{i=1}^{N}\sum_{j=1}^{N}||\boldsymbol{\theta}_{i,j}^{t+1}||^2].
\end{aligned}
\label{asy_theorem1_24}
\end{equation}
\end{footnotesize}

Let $\eta \le \{\frac{(1 - \rho)}{288 L N k_1 \tau}, \frac{\sqrt{1 - \rho}}{24 L \sqrt{N k_1 \tau}}, \frac{\sqrt{1 - \rho}}{24L \sqrt{2(N + M)(T_1 + T)}}, \frac{(1 - \rho)^2}{640L^2N},$ 
\noindent $\frac{1}{10 NL}, \frac{(1-\rho)^{\frac{3}{2}}}{40L}, \frac{1}{4L\sqrt{M}}\}$, $p_2 = \frac{8}{\eta}(\frac{2}{N} + \tau k_1 + \frac{3(N+M)}{32})$, $p_3 = (256L^2\tau k_1 + 32L^2 + 24L^2(N+M))8 \eta$, $p_4 = (\frac{96L^2\tau k_1 + 9L^2(N+M)}{1-\rho} + 32L^2)8\eta$, $p_5 = \frac{64\tau k_1 L + 6L(N+M)}{1-\rho} + \frac{16}{N} $, $p_6 = p_7 =\frac{3}{\eta}$, and set $p = \max \{p_2, p_3, p_4, p_5, p_6, p_7\}$, we have:
\begin{footnotesize}
\begin{equation}
\begin{aligned}
C_i'\le p C_i, i = 2,3,4,5,6,7.
\end{aligned}
\label{asy_theorem1_37}
\end{equation}
\end{footnotesize}

Multiply both sides of Eq.(\ref{asy_theorem1_24}) by $p$, sum Eq.(\ref{asy_theorem1_24}) and Eq.(\ref{asy_theorem1_26}) from $t = T_1 +2 \cdots T_1+T$ and divide by $T-1$. According to Eq.(\ref{asy_theorem1_37}) we have
\begin{footnotesize}
\begin{equation}
\begin{aligned}
\begin{aligned}
& \frac{1}{T-1} \sum_{t = T_1 +2}^{T_1 +T} \mathbb{E}[{\Psi}^t]\\
\le & \frac{p}{T-1}(\frac{F^{T_1+2}- \underline{L}}{N^2M}  +\frac{c_1^{1} \alpha_{1}}{2N} + \frac{c_2^{1} \alpha_{2}}{2M} +\frac{4}{N\eta} (\frac{c_{1}^{0}}{c_{1}^{1}} + \frac{c_{1}^{1}}{c_{1}^{2}})\alpha_{1} 
+ \frac{4}{M\eta} (\frac{c_{2}^{0}}{c_{2}^{1}}\\
+ & \frac{c_{2}^{1}}{c_{2}^{2}}) \alpha_{2} +  5(c_1^{1})^2\alpha_1 + 5(c_2^{1})^2\alpha_2 + \frac{6}{N\eta}\sigma_{1}^{2} + \frac{6}{M\eta} {\sigma_{2}}^{2} + \frac{c_{1}^{2}}{2N}\alpha_{1}+\frac{c_{2}^{2}}{2M}\alpha_{2}) \\
= & \mathcal{O}(\frac{1}{T-1}),\\
\end{aligned}
\end{aligned}
\label{asy_theorem1_38}
\end{equation}
\end{footnotesize}

\noindent where $\sigma_1 = \max \{||\lambda_1 - \lambda_2||\}$, $\sigma_2 = \max \{||\boldsymbol{\theta}_1 - \boldsymbol{\theta}_2||\}$, $\underline{L} = \min L_p(\{\bar{\mathbf{x}}^{t}\},\{\bar{\mathbf{y}}^{t}\}, \{\lambda_{i,l}^{t}\},\{\boldsymbol{\theta}_{i,j}^{t}\})$ satisfies $\forall t \ge T_1 +2$, $F^t \ge \underline{L}-\frac{4}{\eta} \frac{c_{1}^{1}}{c_{1}^{2}} NM \alpha_{1}-\frac{4}{\eta} \frac{c_{2}^{1}}{c_{2}^{2}} N^2 \alpha_{2}-\frac{6}{\eta} NM \sigma_{1}^{2}-\frac{6}{ \eta} N^2{\sigma_{2}}^{2}-\frac{c_{1}^{2}}{2} NM \alpha_{1}-\frac{c_{2}^{2}}{2} N^2 \alpha_{2}$. 
\end{proof}

\subsection{Proof of Theorem 2}
\begin{proof}
In each iteration, after updating local parameters according to Eq.(17)-Eq.(18), each agent broadcasts the local variables $\mathbf{x}_{i} \in \mathbb{R}^n$ and $\mathbf{y}_{i}\in \mathbb{R}^m$ to neighbors. Then after updating dual variables according to Eq.(19)-Eq.(20), each active agent $i$ broadcasts ${\lambda}_{i,l} \in \mathbb{R}^1, l \in |\mathcal{P}|$ and $\boldsymbol{\theta}_{i,j} \in \mathbb{R}^n, j \in \mathcal{N}_i$ to neighbors. In summary, the communication complexity of the $t^{th}$ iteration can be calculated as follows:

\begin{footnotesize}
\begin{equation}
C_1^t = \underbrace{32d^t (N(m+n)}_{\text{Eq.(17), Eq.(18)}}+ \underbrace{\sum_{i=1}^Np_i(|\mathcal{P}_i^t|+nd^t))}_{\text{Eq.(19), Eq.(20)}},
\label{eq_cc_1}
\end{equation}
\end{footnotesize}

\noindent where $d^t = \frac{1}{N}\sum_{i=1}^{N}\sum_{j=1}^{N}\mathbf{W}^t$ denotes the average degree across all nodes.

Before the $T_1^{th}$ iteration, Argus updates cutting planes every $\iota$ iterations. In each time to update cutting planes, Argus first estimates the lower-level problem as Eq.(5)-Eq.(6) and each agent broadcasts ${\mathbf{y}'_{i}}^{(k+1)} \in \mathbb{R}^m$ and $\boldsymbol{\varphi}_{ij}^{(k+1)} \in \mathbb{R}^m$ to neighbors. Then Argus updates cutting planes as Eq.(23) and agents broadcast $\boldsymbol{{\mathcal{P}}}_i^{t+1}$ and $\{\lambda_{i}^{t+1}\}$ to neighbors. The communication complexity of updating cutting planes is:
\begin{footnotesize}
\begin{equation}
% C_2^t = 32Nd^tK(m+mN)+32Nd^t(N(n+m) + 1)
C_2 = 32\sum_{t\in \mathcal{T}}(Nd^t(\underbrace{K(m+m d^t)}_{\text{Eq.(5), Eq.(6)}} + \underbrace{(d^t(n+m) + 1)}_{\text{$\boldsymbol{{\mathcal{P}}}_i^{t+1}$, $\{\lambda_{i}^{t+1}\}$ }})),
\label{eq_cc_2}
\end{equation}
\end{footnotesize}
where $\mathcal{T} = \{1, \cdots, \left\lfloor \frac{T_1}{\iota}\right\rfloor\} \cdot \iota $.

According to Eq.(\ref{eq_cc_1}) and Eq.(\ref{eq_cc_2}), we can obtain that the overall communication complexity of Argus is:
\begin{footnotesize}
\begin{equation}
\mathcal{O}(\sum_{t=1}^TC_1^t + C_2).
\label{eq_cc_3}
\end{equation}
\end{footnotesize}
\end{proof}

\subsection{Proof of Theorem 3}
\begin{proof}
We first calculate the FLOPs of updating local variables. In Eq.(15) and Eq.(16), each agent aggregates the variables of neighbors. In Eq.(17), active agents perform the proximal operator after a step of gradient descent. According to Assumption 1(b), we assume the proximal operator can be accomplished in one step. In Eq.(18), active agents perform a step of gradient descent. Besides, in Eq.(19) and Eq.(20), active agents perform gradient ascent. In summary, the computational complexity of the $t^{th}$ iteration is:

\begin{footnotesize}
\begin{equation}
\begin{aligned}
C_{P_{1}}^{t} & = \underbrace{\mathcal{O}\left(N d^{t}(n+m)\right)}_{\text{Eq.(15), Eq.(16)}} + \underbrace{\mathcal{O}\left(N\left|\mathcal{P}_{i}^{t}\right| d^{t}(n+m)\right)}_{\text{Eq.(17)}} \\
&+ \underbrace{\mathcal{O}\left(N\left|\mathcal{P}_{i}^{t}\right|^{2} d^{t}(n+m)\right)}_{\text{Eq.(18)}} + \underbrace{\mathcal{O}\left(N d^{t^{2}} n\right)}_{\text{Eq.(19), Eq.(20)}}\\
& = \mathcal{O}\left(N\left|\mathcal{P}_{i}^{t}\right|^{2} d^{t}(n+m)\right) + \mathcal{O}\left(N d^{t^{2}} n\right),
\label{eq_coc_1}
\end{aligned}
\end{equation}
\end{footnotesize}

\noindent where $d^t = \frac{1}{N}\sum_{i=1}^{N}\sum_{j=1}^{N}\mathbf{W}^t$ denotes the average degree across all nodes.

Then we calculate the FLOPs of updating cutting planes. Before the $T_1^{th}$ iteration, Argus updates cutting planes every $\iota$ iterations. When estimating the lower-level problem according to Eq.(5)-Eq.(6), Argus utilizes the proximal gradient descent to update $\mathbf{y}_i$ and utilizes the gradient ascent to update the dual variable $\boldsymbol{\varphi}$. After that, Argus calculates the parameters of the new cutting plane according to Eq.(24)-Eq.(26). In summary, the computational complexity of updating cutting planes is:
\begin{footnotesize}
\begin{equation}
C_{P_{2}} = \sum_{t\in \mathcal{T}}(\mathcal{O}(Nd^t(n+m) + NmK)).
\label{eq_coc_2}
\end{equation}
\end{footnotesize}

According to Eq.(\ref{eq_coc_1}) and Eq.(\ref{eq_coc_2}), the overall computational complexity of Argus is:
\begin{equation}
\mathcal{O}(\sum_{t=1}^TC_{P_{1}}^t + C_{P_{2}}).
\label{eq_coc_3}
\end{equation}
\end{proof}

\end{document}